\documentclass[lettersize,journal]{IEEEtran}
\usepackage{amsmath,amsfonts,amssymb}

\usepackage{hyperref}
\usepackage{bbm}
\usepackage[capitalize]{cleveref}
\crefname{section}{Sec.}{Secs.}
\Crefname{section}{Section}{Sections}
\Crefname{table}{Table}{Tables}
\crefname{table}{Tab.}{Tabs.}
\usepackage[draft]{pgf}

\usepackage{xspace}
\makeatletter
\DeclareRobustCommand\onedot{\futurelet\@let@token\@onedot}
\def\@onedot{\ifx\@let@token.\else.\null\fi\xspace}
\def\eg{\emph{e.g}\onedot}

\def\etal{\emph{et al}\onedot}
\makeatother
\usepackage{stfloats}
\usepackage{multirow}
\usepackage{cuted}
\usepackage{capt-of}
\usepackage{float}%
\usepackage{caption}%
\usepackage{graphicx}
\usepackage{xcolor}
\newcommand{\first}[1]{\textcolor[rgb]{0.8,0,0}{{#1}}}
\newcommand{\second}[1]{\textcolor[rgb]{0,0,0.8}{{#1}}}

\usepackage{booktabs}

\usepackage{wrapfig}

\usepackage{algorithmic}
\usepackage{algorithm}
\usepackage{array}
\usepackage[caption=false,font=normalsize,labelfont=sf,textfont=sf]{subfig}
\usepackage{textcomp}
\usepackage{stfloats}
\usepackage{url}
\usepackage{verbatim}
\usepackage{graphicx}
\usepackage{cite}
\usepackage{multirow}

\begin{document}

\title{Personalized Image Filter: \\Mastering Your Photographic Style}

\author{Chengxuan~Zhu, Shuchen~Weng, Jiacong~Fang, Peixuan~Zhang, Si~Li, \\Chao~Xu,~\IEEEmembership{Member,~IEEE}, and~Boxin~Shi,~\IEEEmembership{Senior~Member,~IEEE}%
  \IEEEcompsocitemizethanks{
  \IEEEcompsocthanksitem Chengxuan~Zhu and Chao Xu are with State Key Laboratory of General Artificial Intelligence, School of Intelligence Science and
Technology, Peking University, Beijing 100871, China.
    \IEEEcompsocthanksitem Shuchen~Weng is with Beijing Academy of Artificial Intelligence, Beijing 100083, China.
    \IEEEcompsocthanksitem Jiacong~Fang and Boxin~Shi (corresponding
author: \protect\url{shiboxin@pku.edu.cn}) are with the State Key Laboratory of Multimedia Information Processing and National Engineering Research Center of Visual Technology, School of Computer Science, Peking University, Beijing 100871, China.
\IEEEcompsocthanksitem Peixuan~Zhang and Si Li are with School of Artificial Intelligence, Beijing University of Posts and Telecommunications, Beijing 100876, China.
    \IEEEcompsocthanksitem This work was supported by National Natural Science Foundation of China (Grant No. 62136001 and 62276007), Beijing Municipal Science \& Technology Commission, Administrative Commission of Zhongguancun Science Park (Grant No. Z241100003524012), and Beijing Natural Science Foundation (Grant No. QY24037).
  }
}
\markboth{IEEE TRANSACTIONS ON PATTERN ANALYSIS AND MACHINE INTELLIGENCE}{}

\maketitle
\begin{abstract}
Photographic style, as a composition of certain photographic concepts, is the charm behind renowned photographers. But learning and transferring photographic style need a profound understanding of how the photo is edited from the unknown original appearance. Previous works either fail to learn meaningful photographic concepts from reference images, or cannot preserve the content of the content image. To tackle these issues, we proposed a Personalized Image Filter (PIF). Based on a pretrained text-to-image diffusion model, the generative prior enables PIF to learn the average appearance of photographic concepts, as well as how to adjust them according to text prompts. PIF then learns the photographic style of reference images with the textual inversion technique, by optimizing the prompts for the photographic concepts. PIF shows outstanding performance in extracting and transferring various kinds of photographic style. Project page: \url{ https://pif.pages.dev/}
\end{abstract}

\begin{figure}[h]
    \centering
    \includegraphics[width=\linewidth]{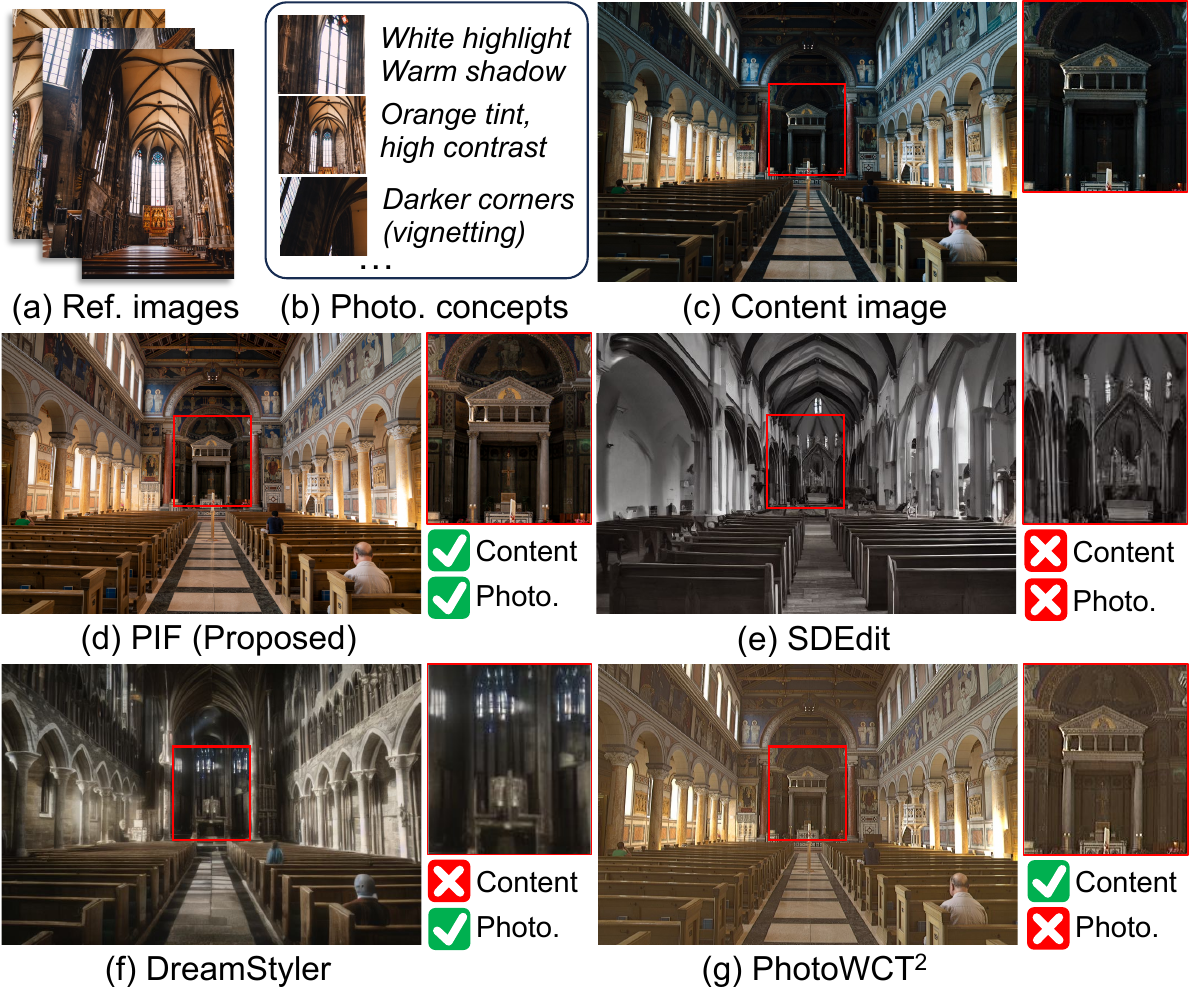}
    \caption{
    Given several (a) reference images, PIF learns a set of (b) photographic (Photo.) concepts, and can render any (c) content image with the learned photographic style from (a), as shown in (d). In comparison, the text-based image editing method, SDEdit~\cite{meng2021sdedit}, and personalization method, DreamStyler~\cite{ahn2024dreamstyler}, change the content completely. The reference-based image retouching method PhotoWCT$^2$~\cite{chiu2022photowct2} only transfers the color distribution instead of the photographic concepts.}
    \label{fig:teaser}
\end{figure}
\section{Introduction}\label{sec:intro}
\begin{figure*}[t]
    \centering
    \subfloat[Sharpness (strong/weak)\label{fig:sharpness}]{
        \includegraphics[width=0.24\linewidth]{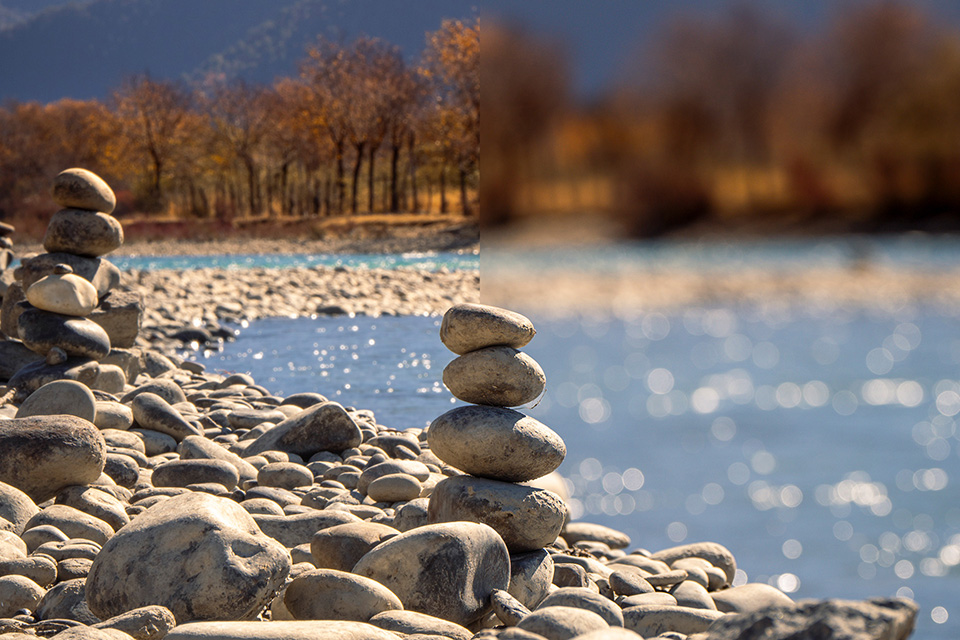}
        }%
    \subfloat[Vignetting (dark/none)]{
        \includegraphics[width=0.24\linewidth]{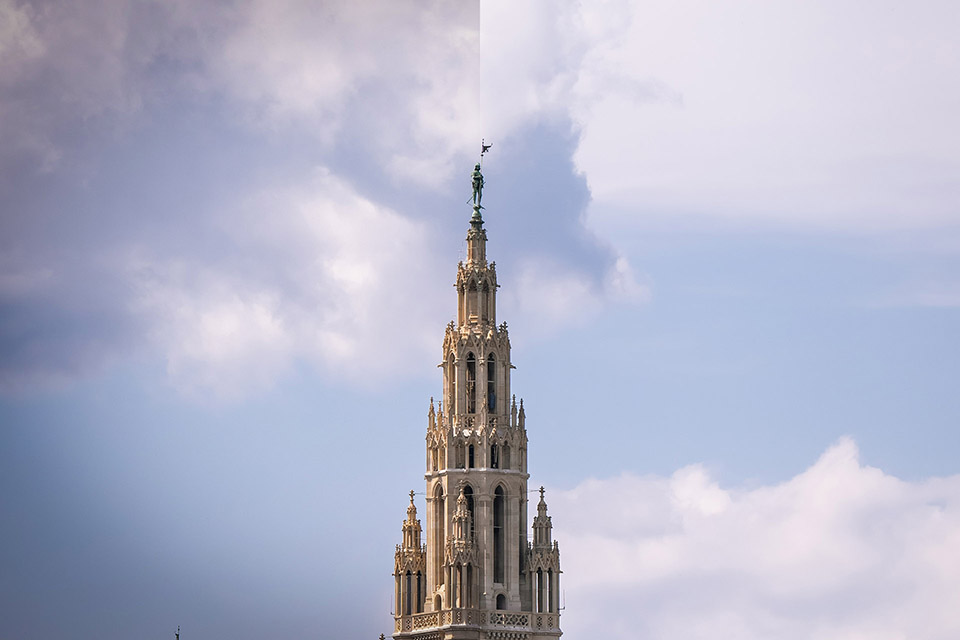}
        }%
    \subfloat[Saturation (high/low)]{
    \includegraphics[width=0.24\linewidth]{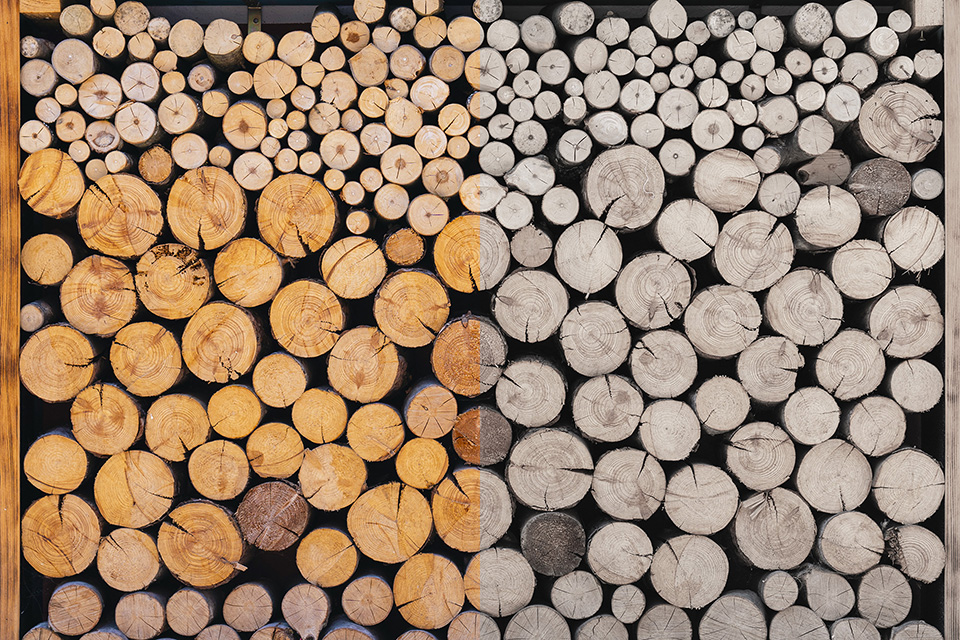}
    }%
    \subfloat[Tint (cyan/purple)]{
    \includegraphics[width=0.24\linewidth]{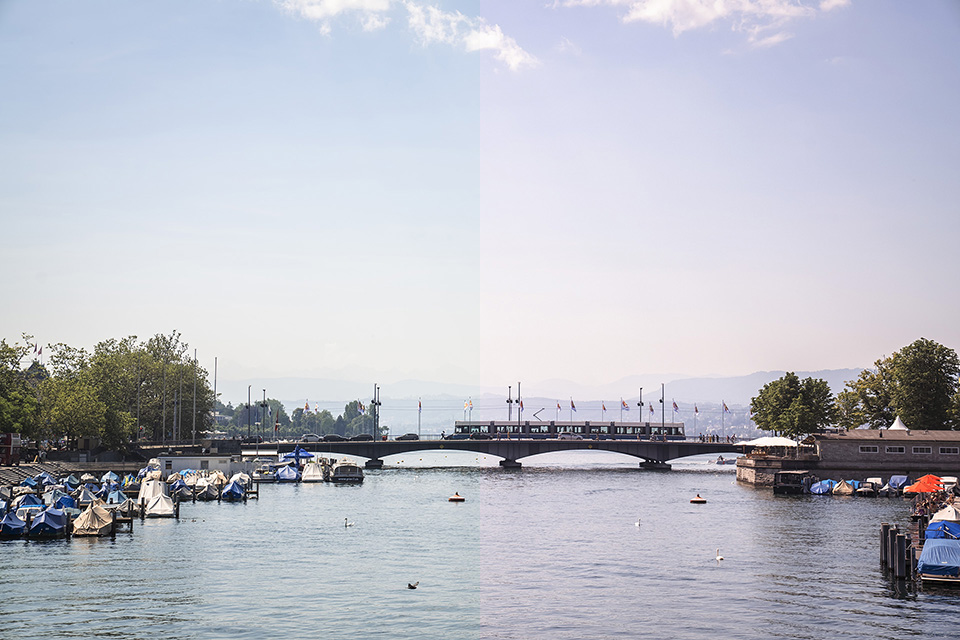}
    }\hfill\\
    \subfloat[Exposure (high/low)]{
    \includegraphics[width=0.24\linewidth]{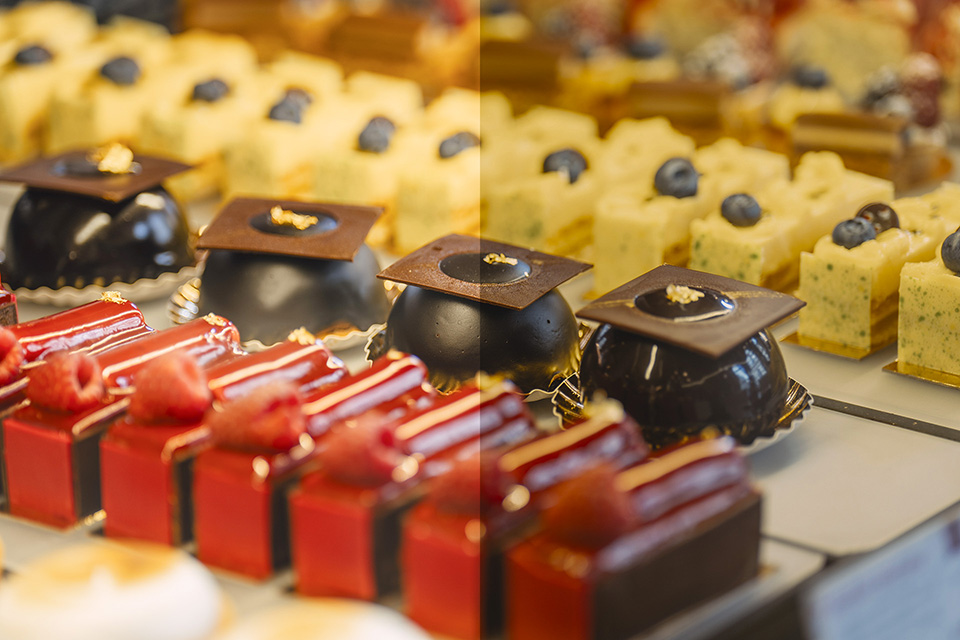}
    }%
    \subfloat[Contrast (high/low)]{
    \includegraphics[width=0.24\linewidth]{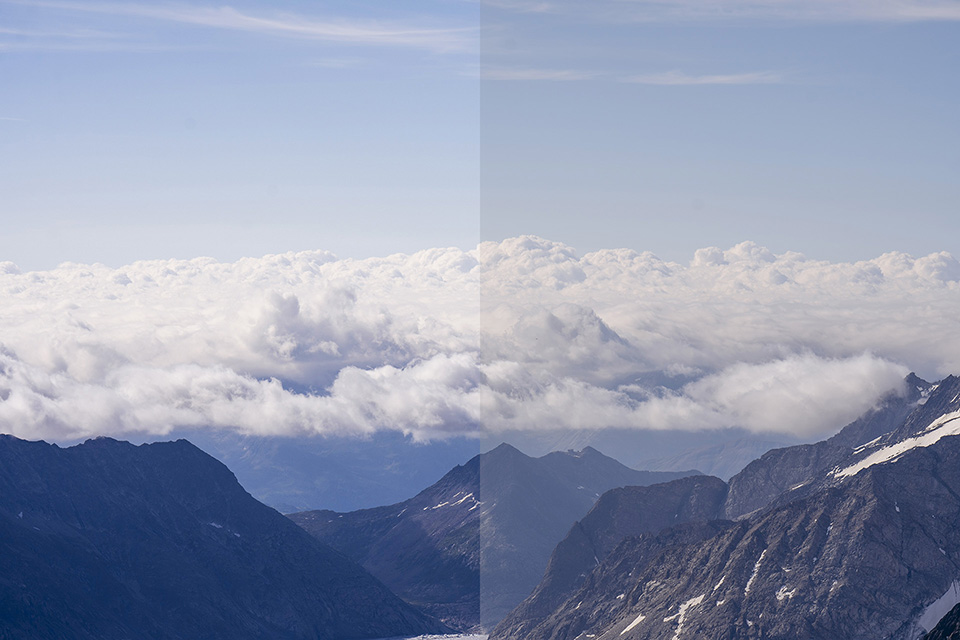}
    }%
    \subfloat[Highlight (gray/cyan)]{
    \includegraphics[width=0.24\linewidth]{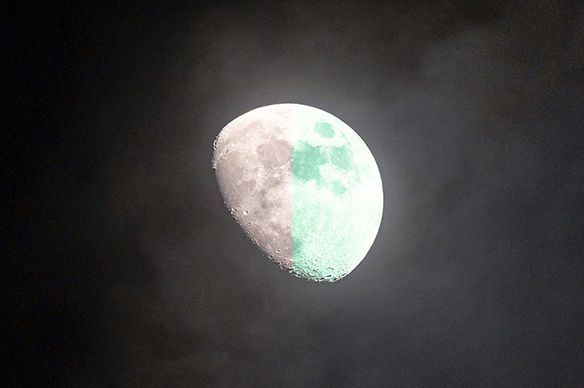}
    }%
    \subfloat[Shadow (gray/blue)]{
    \includegraphics[width=0.24\linewidth]{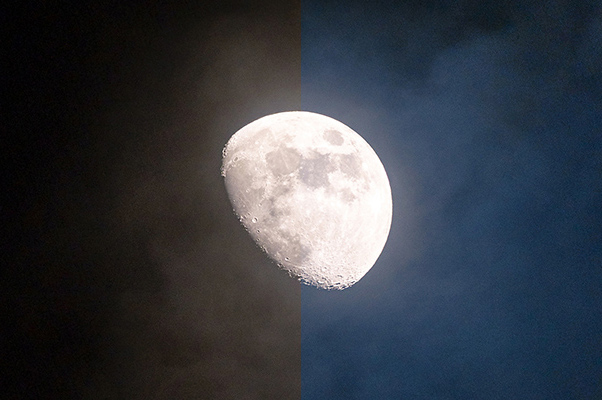}
    }\hfill
    \caption{The eight types of photographic concepts employed in this paper, which are used to decompose the photographic style. The left and right halves of each image correspond to different appearances of the concepts, as described in the captions below.}
    \label{fig:visualize-concept}
\end{figure*}

\begin{quote}
\emph{``A picture is the expression of an impression.''} 

\hfill --- Ernst Haas
\end{quote}
The words of esteemed photojournalist Ernst Haas convey a fundamental principle of photography: Photography is not only about the contents, but the embedded photographic style, such as those listed in \cref{fig:teaser}(b).
Composed of a combination of \textit{photographic concepts} (\eg, exposure, contrast and tint, as illustrated in \cref{fig:visualize-concept}), \textit{photographic style} distinguishes the photos from the content, namely the objects' color and shape. For a given photo, the photographic style is a specific combination of photographic concepts that deviate from the average ones, and conveys the photographer's preferences and aesthetic ideology. 

In this paper, we propose the new task of learning the photographic style from images. It is realized by the design of a \textit{Personalized Image Filter} (PIF), to endow non-professional users with the photographic styles of any professional. Given a set of reference images (\cref{fig:teaser}(a)), PIF aims to learn the photographic style by optimizing the photographic concepts, such as the concepts in \cref{fig:teaser}(b). After that, it should be able to apply the learned photographic style to any given content image (\cref{fig:teaser}(c)), for a result with the photographic style similar to reference images, and keep the content the same as in the content image, as shown in \cref{fig:teaser}(d).
Though the objective sounds familiar, it needs a fine-grained understanding about each photographic concept to know the personalized photographic style. To render the photographic style while keeping the content intact poses another challenge.

Related methods can only address a fraction of the issue:
\textit{(i)} \textbf{Text-based image editing}~\cite{labs2025flux1kontextflowmatching,meng2021sdedit,brooks2023instructpix2pix,kwon2022clipstyler},
though having more control capability thanks to text conditioning, the content is prone to be changed~\cite{meng2021sdedit,brooks2023instructpix2pix}, as observed \cref{fig:teaser}(e). The photographic concepts may also be misinterpreted or insufficiently described~\cite{labs2025flux1kontextflowmatching,kwon2022clipstyler}.
\textit{(ii)} \textbf{Text-to-image personalization}~\cite{gal2022image,avrahami2023break} can optimize a word embedding denoting the concept presented in the image without explicit text description~\cite{ahn2024dreamstyler,gal2022image}, but the diffusion prior tends to modify the content, as shown in \cref{fig:teaser}(f).
\textit{(iii)} \textbf{Image retouching with reference} is another closely related field, which aims at transferring the style of a reference image to a content image. However, as \cref{fig:teaser}(g) suggests, they tend to overfit the color and lightness distribution, which is caused by missing an anchor on the average appearance of the content.

In our design, to preserve the content while transferring the professional photographic style, we tailor a brand new diffusion-based framework, consisting of a residual one-step diffusion and a photographic concept perturbation, to replace the diffusion and denoising process of diffusion models.
To learn the photographic style, PIF first finetunes an off-the-shelf text-to-image model to learn the average style and the binding between text instructions and professional photographic concepts from the diffusion priors. It is designed to inverse the process of a photographic concept perturbation similar to denoising, and the denoising process is also modified to a residual denoising scheme, to retain the original structure and content.
Based on the average style, PIF optimizes multiple tokens for the various professional photographic concepts a random combination training strategy to bind the photographic concepts and the pseudo words together, to master the photographic concepts separately.

In brief, the contributions of this work are:
\begin{itemize}
    \item a white-box photographic style transfer framework, based on diffusion prior to understand and render the photographic style the editing instructions about photographic concepts given by the text prompts,
    \item a novel detail-preserving denoising paradigm that predicts the residual from the content image, bypassing the problem of fidelity caused by diffusion models, and
    \item a photographic style transfer method that accurately renders the photographic concepts from several reference images, utilizing a random combination training strategy and the textual inversion technique.
\end{itemize}
\section{Related Work}\label{sec:related}

We discuss three aspects of related work: \textit{(A)} text-based image editing, where text description is used to guide the editing of images; \textit{(B)} text-to-image personalization, which aims to adapt models to recognize and generate~\cite{nitzan2022mystyle} or discriminate~\cite{cohen2022my} concepts from a limited set of unique category images; \textit{(C)} image retouching with reference that renders the content image in the style of the reference images. 
\subsection{Text-Based Image Editing}\label{sec:related:TIEditing}
Text-based image editing methods have developed in various directions.
Conditional generation~\cite{wang2018high,odena2017conditional,li2019object,xie2023boxdiff,li2018closed,meng2021sdedit,brooks2023instructpix2pix} encodes text as an additional condition, but the generated contents tend to differ from the input image, and the desired photographic concepts can hardly be understood by pre-trained text encoding models.
Inversion-based~\cite{nguyen2023visual,roich2022pivotal,tov2021designing,richardson2021encoding,delbracio2023inversion} methods aim to optimize a noise latent that can be used to generate the content image, and thus modify the denoising process slightly to generate desired images, but it is unknown how to modify the latent in a photographic way.
Finetuning~\cite{fei2023generative,zhang2023towards} or new architectural design~\cite{qwenimage,bagel} can optimize the foundation model to follow the prompts more faithfully, but requires a prohibitively large number of paired images and texts.
Sampling-based methods~\cite{voynov2023p+,ding2021cogview,huang2021unifying,gafni2022make} aim to sample the output with the guidance of the input text, but fall short in preserving the image content. 
Some recent methods are able to perform subject-preserving editing, handling versatile tasks~\cite{labs2025flux1kontextflowmatching,lee2024cliptone}, but they struggle to edit the photographic style following the instructions.

We propose a one-step residual denoising scheme, estimating the difference between the input and the target through the difference between the desired prompts and null prompts. The diffusion process is also substituted with the photographic concept perturbation, so that the original structure is preserved both in degradation and restoration.

\subsection{Text-to-Image Personalization}\label{sec:personalization}
Diffusion models are among the most preferred schemes for text-to-image personalization tasks, thanks to the powerful diffusion prior acquired on large-scale pretraining. Recently, some methods are able to edit images according to text instructions~\cite{meng2021sdedit,labs2025flux1kontextflowmatching,lee2024cliptone}, but the prompts may not reflect the desired style accurately, and may also be misinterpreted by the text-based image editing models.

To save the trouble of describing reference images with texts, image personalization methods have come up with the idea of finetuning the model on the reference images~\cite{ruiz2023dreambooth,kumari2023multi,wei2023elite,avrahami2023break,ye2023ip}, optimizing new word embeddings for the desired contents~\cite{gal2022image,wei2023elite,avrahami2023break,chen2023anydoor}, adding more conditioning to the noise prediction network~\cite{voynov2023p+,Zhang_2023_ICCV}, and employing prompt-to-prompt techniques~\cite{hertz2022prompt,brooks2023instructpix2pix} to steer the sampling process.
However, these personalization methods can only memorize visually apparent and separated concepts~\cite{avrahami2023break}, if there are multiple concepts, while the photographic concepts tend to intertwine with each other.
In this work, we aim to personalize the text descriptions for the photographic concepts. With a fine-tuned model that is aware of the average appearance of photographic concepts, PIF is able to optimize the pseudo word embeddings to match the photographic concepts in the reference images, with a random combination training strategy.

\subsection{Image Retouching with Reference}
Image retouching aims to improve the quality of the photos, while keeping the content unchanged. Due to the subjective nature of beauty, users often wish to guide the image retouching process with a reference image. Early works mainly focus on learning the color distribution in the reference image~\cite{reinhard2001color}, while ignoring the potential color discrepancy. With machine learning, recent methods perform more effectively in extracting the photographic style from the reference image, such as Gram matrix~\cite{gatys2016image}, the whitening and coloring transformation~\cite{li2017universal,yoo2019photorealistic,hong2021domain}, the Wavelet Corrected Transfer~\cite{li2018closed} and attention mechanisms~\cite{park2019arbitrary,wu2021styleformer,StyTr2}. 
However, none of these methods are designed for specific photographic concepts, and thus they tend to overfit on the overall color of the reference image. Moreover, paired dataset is often applied to train the mapping, which is often unavailable in the task of personalizing image filters.
As diffusion models are pretrained on huge amount of data, we aim to leverage the learned average style as an anchor, and learns the personalized photographic style on top of that.

\section{Problem Formulation}\label{sec:definition}

We first formulate the task of photographic style transfer, and then introduce the proposed Personalized Image Filter (PIF) to solve the task. 
The photographic style, by definition, is a composition of multiple photographic concepts. By breaking down how photographers adjust them, as illustrated in \cref{fig:visualize-concept}, the photographic style can be mastered at a finer granularity. 

For a set of reference images $\{I_\text{R}^{(i)}\}_{i=1}^N$, the model should extract a set of photographic concepts $P_\text{U} = \{p_j\}_{j=1}^M$ embedded in the reference images, and then impose the learned photographic style on any given content image $I_\text{C}$, yielding $I_\text{P}=f_{P_\text{U}}(I_\text{C})$. $I_\text{P}$ is expected to inherit the contents from $I_\text{C}$, and all the photographic concepts in $P_\text{U}$.

Each photographic concept, by definition, corresponds to a weight function $W_j(\cdot)$ and an adjustment function $f_j(\cdot,\xi)$, aligned with previous methods~\cite{wang2022neural,ho2021deep}. The weight function $W_j(I)\in[0,1]^{(h,w)}$ indicates the strength of the concept, while $\xi\in[-1,1]$ stands for the strength of the adjustment, and the larger $|\xi|$, the more apparent the adjustment. For example, the well-known concept of ``exposure'' should correspond to the weight function of the average intensity of the image, while its adjustment function to adjusting the global intensity by multiplying a global factor. 

To limit the scope, in this paper, we include the following photographic concepts, characterized by their weight functions and adjustment functions. They are given the concept name of $p_1,\cdots, p_8$ for the convenience of notation. 
$\nabla I$ stands for the gradient of the pixel intensity. $I_\text{H}$, $I_\text{S}$, and  $I_\text{V}$ are the hue, saturation, and value channels of $I$.

\textit{(i)} \textbf{Lens-related effects}: sharpness and vignetting. Sharpness ($p_1$, \cref{fig:visualize-concept}(a)) is the clearness of high frequency details, while vignetting ($p_2$, \cref{fig:visualize-concept}(b)) is the typically darker (or brighter) corners in the image, given by
\begin{equation}
    f_1(I, \xi) = (\xi+1)\cdot I - \xi \cdot I_\text{blur}, 
\end{equation}
\begin{equation}
    f_2(I,\xi)_{ik} = I_{ik} \cdot ( 1+\xi\cdot W_2(I) ),
\end{equation}
where $I_\text{blur}$ is a gaussian-blurred version of $I$, parameterized by the kernel size of $11$. The weight functions are defined as
\begin{equation}
W_1(I) = |\nabla I-\overline{\nabla I}|, 
    W_2(I)_{ik} = \frac{(2i-h)^2\!+\!(2k-w)^2}{w^2+h^2}.
\end{equation}

\textit{(ii)} \textbf{Global color}: saturation and tint. Saturation ($p_3$, \cref{fig:visualize-concept}(c)) is the vividness of the colors, and tint ($p_4$, \cref{fig:visualize-concept}(d)) is the global color bias in the image, caused by the configuration of white balance. They are defined in the HSV color space as
\begin{equation}
    f_3(I, \xi)_\text{S} = (\xi+1) \cdot I_\text{S}, 
\end{equation}
\begin{equation}
    f_4(I, \xi)_\text{H} = (1-|\xi|) \cdot I_\text{H}+|\xi| \cdot h(\xi),
\end{equation}
where $h(\xi)$ is a random hue conditioned on $\xi$.
\begin{equation}
    W_3(I) = |I_\text{S}-\overline{I_\text{S}}|,\quad W_4(I) =\textbf{1}.
\end{equation}
For simplicity of notation, we split the parameter $\xi$ into two part, $\xi^{(1)},\xi^{(2)}\in[0,1]$, controlling the intensity and the hue of the color bias.

\textit{(iii)} \textbf{Illumination responses}: exposure and contrast. Exposure ($p_5$, \cref{fig:visualize-concept}(e)) indicates the mean brightness of the image, while contrast ($p_6$, \cref{fig:visualize-concept}(f)) describes the difference between the brightest and darkest areas in the image defined as
\begin{equation}
    f_5(I, \xi) = I \cdot (1 + \xi), 
\end{equation}
\begin{equation}
f_6(I, \xi) = \bar{I} + (\xi + 1) \cdot (I - \bar{I}),
\end{equation}
and the weight function is given by
\begin{equation}
    W_5(I) = |I-\overline{I}|,
\end{equation}
\begin{equation}
    W_6(I) = (I-\frac{2I_\text{max}+I_\text{min}}{3})\odot (I-\frac{I_\text{max}+2I_\text{min}}{3}).
\end{equation}

\textit{(iv)} \textbf{Histogram-related area}: highlight ($p_7$, \cref{fig:visualize-concept}(g)) and shadow ($p_8$, \cref{fig:visualize-concept}(h)) indicate the color of bright area and dark area in the image respectively, which professional photographers often adjust to create a unique charm in the photo. They are given by
\begin{equation}
    f_7(I, \xi) = (1-W_7(I))\odot I +\mathbf{c}_\text{h}(\xi) W_7(I),
\end{equation}
\begin{equation}
    f_8(I, \xi) = (1-W_8(I))\odot I +\mathbf{c}_\text{s}(\xi) W_8(I),
\end{equation}
\begin{equation}
    W_7(I) = \max(0, \frac{I-\tau_\text{h}}{1-\tau_\text{h}}),\quad W_8(I) = \max(0, \frac{\tau_\text{s}-I}{\tau_\text{s}}),
\end{equation}
where \(\tau_\text{h}\) and \(\tau_\text{s}\) are thresholds for bright area and dark area respectively (\eg, $0.7$ and $0.3$ in normalized intensity), while $\mathbf{c}_\text{h}(\xi), \mathbf{c}_\text{s}(\xi) \in[0,1]^3$ are RGB color vectors controlled by $\xi$.

The mentioned photographic concepts are some of the most commonly used ones by photographers when developing the photos from raw image files. Therefore, PIF aims to decompose the photographic style into these eight types of photographic concepts.
Note that PIF can be expanded to include more photographic concepts, as long as they are defined mathematically.

\section{Method}\label{sec:method}
The training of PIF is a two-stage pipeline. First, to make PIF aware of the average photographic concepts, and to preserve the content while modifying the photographic concepts, an off-the-shelf SD v1.5 text-to-image model~\cite{sg161222} is finetuned for a novel design of residual one-step diffusion. The first stage is shown in \cref{fig:pipeline}, and it only needs to be trained once.
With the trained and fixed residual one-step diffusion backbone, in the second stage PIF learns the photographic concepts from the reference images with a multi-concept inversion technique, as shown in \cref{fig:pipeline_stage3}. The second stage needs to be optimized for each set of reference images before transferring the professional photographic style to arbitrary images.

We first introduce some preliminary knowledge about the diffusion models in \cref{sec:prelim_diff}.
Then we propose the residual denoising scheme with only one inference step in \cref{sec:one_step}, which uses the perturbation of photographic concepts as the forward diffusion process, enabling the noise prediction model to both preserve the image content and to edit the photographic concepts given instructions. 
With the priors of average photographic style learned as an anchor, PIF is able to learn the photographic style by optimizing the pseudo words describing the photographic concepts, as detailed in \cref{sec:PCLO}.

\subsection{Preliminaries for Diffusion Models}\label{sec:prelim_diff}
Diffusion models are pre-trained to remove noise from the input latent~\cite{ho2020denoising}, by iteratively denoising the latent with a noise prediction network $\epsilon_\theta$, defined by the forward diffusion process of
\begin{equation}
    z_t = \sqrt{\bar{\alpha}_t}z_0 + \sqrt{1-\bar{\alpha}_t}\epsilon, \quad \epsilon\sim\mathcal{N}(\mathbf{0},\mathbf{I}),
\end{equation}
and the denoise sampling process of
\begin{equation}
\hat{z}_{t-1}= \frac{1}{\sqrt{\alpha_t}}(z_t - \frac{\beta_t}{\sqrt{1-\bar{\alpha}_t}} \epsilon_\theta(z_t; y_\text{txt})) + \sigma_t \psi,
\label{eq:denoise_iter}\end{equation}
where $\psi\sim \mathcal{N}(\mathbf{0},\mathbf{I})$ is the noise in the sampling process. $\beta_t$ is the predetermined variance of the diffusion process at timestep $t$, such that $\sigma_t^2 = \beta_t$, $\alpha_t=1-\beta_t$, and $\bar{\alpha}_t=\prod\alpha_{i\le t}$. The text prompt $y_\text{txt}$ serves as the condition. 

In inference and finetuning phase, the iterative sampling process is often simplified, since the fully denoised latent can be coarsely predicted by
\begin{equation}
    \hat{z}_0 = \frac{z_t - \beta_t \cdot  \epsilon_\theta(z_t; y_\text{txt})}{\alpha_t}.
\label{eq:diffusion}
\end{equation}

As found by previous works, diffusion models tend to perform better in the time-steps they are trained on~\cite{hang2023minsnr}, implying the possibility of a one-step diffusion model, aligned with previous practices~\cite{parmar2024img2imgturbo,osediff,zhu2025bokehdiff} which are trained on a particular timestep.
Even without adding noise, diffusion models can also be finetuned to match the desired data distribution~\cite{osediff,zhu2025bokehdiff}, essentially learning the mapping from $I'$ to $I$ following the text prompts $y_\text{txt}$ while keeping the structure and semantics, given by
\begin{equation}
    \hat{I}(I',\theta,T,y_\text{txt}) = \mathcal{D}\left(\frac{\mathcal{E}(I')-\beta_T\cdot\epsilon_\theta(\mathcal{E}(I');y_\text{txt})}{\alpha_T}\right).
\end{equation}

With these designs, and large-scale pretraining on paired data (image with captions), diffusion models learn to transform random noise to the distribution of meaningful data. However, when editing photographic style, it is crucial to retain the high frequency texture and details of the content image $I'$, which demands a high reconstruction ability. 

As the community have found that latent diffusion models face the dilemma between generation and reconstruction~\cite{yao2025reconstruction,esser2024scaling,gupta2024photorealistic}, the first problem of photographic style transfer is posed by the distortion of diffusion models. 
Though researchers have come up with more accurate autoencoders~\cite{sd3,flux}, models that are finetuned on editing data~\cite{labs2025flux1kontextflowmatching}, or workarounds with higher resolution~\cite{yu2024supir}, the textures and details are still prone to change in the diffusion process.

Recently, it is found that diffusion models do not necessarily involve adding and removing noise: Down-sampling, masking, blurring, and other image manipulation operations can also be the degradation choice~\cite{bansal2024cold,zhu2025bokehdiff,osediff}, as long as the corresponding restoration is defined. This motivates us to consider a diffusion model that is finetuned to learn the adjustment of photographic style, while the content is kept intact both in degradation and restoration.

\subsection{
Residual One-Step Diffusion}\label{sec:one_step}
\begin{figure}[t]
    \centering
    \includegraphics[width=\linewidth]{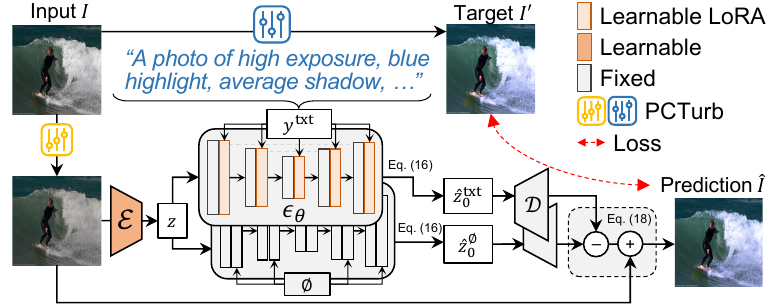}
    \caption{The pipeline of the residual one-step diffusion process, introduced in \cref{sec:one_step}. The photographic concepts in the input image $I$ are adjusted with the parameters of $\{\xi_j'\}$, while generating a description $y_\text{txt}$ of the target image $I'$. Given another adjusted image (parameterized by $\{\xi_j\}$), the noise prediction network is finetuned to predict the noise towards $I'$ by \cref{eq:diffusion}, to learn the average appearance and adjustment of the photographic concepts. To preserve the content, a residual denoising scheme is applied, anchoring the null text prompts $\varnothing$, as defined by \cref{eq:res_denoise}.
    }
    \label{fig:pipeline}
\end{figure}
To preserve the content from the input image $I'$ while receiving photographic style control from the proposed model, we propose a novel residual one-step diffusion paradigm, as illustrated in \cref{fig:pipeline}.
The residual term from the content image is computed from the difference between the prediction conditioned on the control instruction $y_\text{txt}$ and a null prompt $\varnothing$ indicating average style, given by
\begin{equation}\label{eq:res_denoise}
    \hat{I}_\text{res}= I'+\hat{I}(I',\theta,T,y_\text{txt})-\hat{I}(I',\theta,T,\varnothing).
\end{equation}

As for the degradation definition, we define a photographic concept perturbation (PCTurb) which only adjust the photographic concepts with the corresponding adjustment functions $f_j(\cdot, \xi)$, defined by
\begin{equation}
    \mathcal{P}(I, \{\xi_j\}) = f_M(f_{M-1}( \cdots (f_1(I, \xi_1),\cdots), \xi_{M-1}), \xi_M). 
\end{equation}
The diffusion model tries to inverse the image transformation of PCTurb, given by \cref{eq:diffusion}. 
As PCTurb does not change the image content completely, the perturbed data distribution is still close to the original data distribution. Therefore, the noise prediction network is expected to easily predict the correct noise for reconstruction, and thus we simplify the traditional iterative denoising process to a one-step diffusion process, formulated as
\begin{equation}
    \hat{I} = \mathcal{D}\left(\frac{\mathcal{E}(\mathcal{P}(I, \{\xi_j\})) - \beta\cdot\epsilon_\theta(\mathcal{P}(I, \{\xi_j\}); y_\text{txt})}{\alpha_T}\right),
\label{eq:hatI}\end{equation}
and the target image for supervision is given by $I' = \mathcal{P}(I, \{\xi'_j\})$, where $\{\xi_j\}$ and $\{\xi_j'\}$ are two different sets of parameters controlling the photographic concepts, and $y_\text{txt}$ corresponds to the PCTurb with the parameters of $\{\xi_j'\}$. As PCTurb is designed to preserve the content of the image, applying different parameters $\{\xi_j\}$ essentially creates a pair of images that only differ in the photographic concepts. 

The one-step diffusion greatly improving the efficiency, while avoiding the cumulative error, following previous discoveries. The training of the one-step diffusion process is shown in \cref{fig:pipeline}.

We finetune the model on the Unsplash~\cite{unsplash} dataset, which has an average photographic style given the law of large numbers, meaning that the unmodified image should have an average photographic concepts.
To endow the noise prediction model of the ability to adjust different photographic concepts, we compose the text prompt as
\begin{equation}
    y_\text{txt}=\text{``A photo of ''}+\bigoplus_{j=1}^M ([v_j']\ \ p_j'),
\label{eq:ytxt_stage1}\end{equation}
where the notation of $\bigoplus$ denotes text concatenation, and the attributes for each concept, $[v_j']$, depends on $\xi_j'$. As the $[v_j']$ are finite pre-determined text attributes, such as the ones in the captions of \cref{fig:visualize-concept}, they should be anchored to a fixed set of adjustment parameters $\{\xi_j'\}$. 
On the other hand, $\{\xi_j\}$ as an augmentation to the input, can be selected arbitrarily, because the diffusion model is expected to map any image after PCTurb into the desired distribution.

We finetune the noise prediction network $\epsilon_\theta$ with LoRA~\cite{hu2021lora} and the encoder $\mathcal{E}$, following recent works~\cite{osediff,zhu2025bokehdiff}, while keeping the decoder $\mathcal{D}$ fixed. This is because the latent distribution, namely how the latents are mapped to the image space, is already learned when pretraining on vast amount of data, and should remain fixed to prevent catastrophic forgetting.

For a pixel-wise content-preserving supervision, it is necessary to apply both PCTurb and the supervision in image space, since the highly compact latent space is not aligned with human perception. 
The loss function for finetuning is designed as a combination of mean square error, perceptual loss, and edge loss, given by
\begin{equation}
    \mathcal{L} = \|I-\hat{I}\|_2 + \lambda_\text{P}\| f_\text{P}(I)-f_\text{P}(\hat{I})\|_2+ \lambda_\text{E}\|\nabla I-\nabla \hat{I}\|_2,
\end{equation}
where $f_\text{P}$ is a pretrained VGG network to extract visual features for perceptual loss~\cite{lpipscite}. 

Finetuned on a subset of the Unsplash dataset~\cite{unsplash}, the noise prediction network learns to retrieve the average photographic style from the generative prior, by predicting the noise for reconstruction. The text prompt is composed by \cref{eq:ytxt_stage1}, where no information about the content of the input image $I$ is given, so as to achieve a separation between the content and the photographic style. During this stage, we finetune low-rank adapters~\cite{hu2021lora} for the noise prediction network and the text encoder with a rank of $8$, and also the encoder to better extract photographic concepts from the reference images.

In this way, we implicitly separates the content from the original photographic style by rethinking the forward process, and condition the new style on the text prompt. The efficacy of the proposed method is validated by experimental results, where we can perform photographic style editing with text prompts, while preserving the contents.

\subsection{Photographic Style Learning}\label{sec:PCLO}

\begin{figure}[t]
    \centering
    \includegraphics[width=\linewidth]{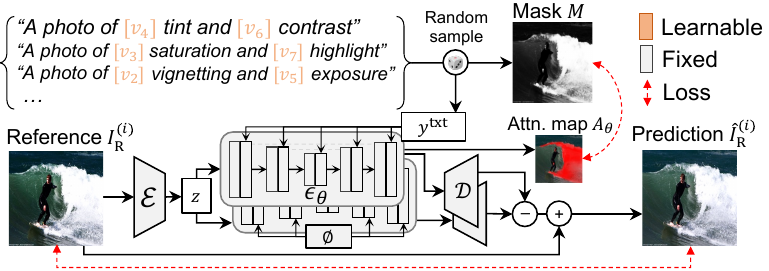}
    \caption{The pipeline of photographic concept learning, described in \cref{sec:PCLO} where photographic concepts are randomly sampled from $P_\text{U}$ (illustrated in the top left bracket), and only the modifiers $[v_j]$ are optimized. The encoder and noise prediction model remain trained and fixed after \cref{sec:one_step}.
    }
    \label{fig:pipeline_stage3}
\end{figure}
Now that the residual one-step diffusion model is able to edit the photographic style while preserving the content, it is then used to learn the photographic style. 
Given the professional photographic concepts defined in \cref{sec:definition}, PIF aims to master the photographic style by optimizing the pseudo attribute words for the concepts through textual inversion.

Though previous works have proved the possibility of imitating the style of images from a few samples with only one new token~\cite{gal2022image}, the optimized single-word embedding is highly abstract and tends to overfit on the given reference images. It also overlooks the mathematical definition of the photographic concepts, as listed in \cref{sec:definition}. 

Therefore, we use multiple pseudo words $\{[v_j]\}$ to learn various photographic concepts in $P_\text{U}$ (defined in \cref{sec:definition}) individually with textual inversion, and then combine all the concepts during inference. 
Following the philosophy that being explicit is better than implicit, we aim to separate different photographic concepts for a more white-box and fine-grained control.
Pseudo words $\{[v_j]\}$ are used to describe the property of concrete photographic concepts $\{p_j\}$ when composing a text prompt. 
The objective of learning photographic style is then turned into learning the embedding vectors of the pseudo words $\{[v_j]\}$ with textual inversion. In this stage, only the text embeddings of pseudo words are optimized.

Inspired by Avrahami \etal~\cite{avrahami2023break}, we propose a random combination training strategy, to learn the text embeddings of $\{[v_j]\}$ of photographic concepts separately. 
At each training step, we randomly sample a subset $P\subseteq P_\text{U}$, and compose the text prompts as
\begin{equation}
    y_\text{txt} = \text{``A photo of ''}+\bigoplus_{p=1}^M \left(\underset{p\notin P}{\mathbbm{1}} \Big([v_j] \  p_j\Big)\right).
\label{eq:text_input}\end{equation} %
$\underset{p\in P}{\mathbbm{1}}(y')$ is an indicator function evaluating to the text $y'$ if and only if $p\in P$, otherwise null. Essentially, the text prompt design encourages only the optimization of the concepts in $P$, while other concepts are set to be average and fixed. This is to prevent the photographic concepts from interfering with each other during optimization, since they are intertwined spatially by definition.

The training objective is to make the predicted reference image $\hat{I}_\text{R}=\mathcal{D}(\hat{z}_0)$ closer to the ground truth $I_\text{R}$ in the region specified by the attention weight functions $W_{p_i}(\cdot)$, and to encourage the model to focus on the regions that are relevant to the photographic concepts specified in the text prompt $y_\text{txt}(P)$, formulated as
\begin{equation}
    \mathcal{L}_{\text{WR}} = \mathbb{E}_{P,I_\text{R}} \left[ \sum_{p\in P} \left\| W_{p}(I)\odot (f_p(I) - f_p(\hat{I}) )\right\|_2 \right],
\label{eq:masked_recon}\end{equation}
which calculates the sum of reconstruction loss of each feature defined by $f_p(\cdot)$ in the region specified by the weight function $W_p(I)$, and is used to supervise the diffusion process.
We further apply a supervision on the cross attention response of the model, which captures how much the model response at each photographic concept, given by the attention loss as
\begin{equation}
    \mathcal{L}_{\text{AT}} = -\mathbb{E}_{P,z^{(i)}} \left\| A_\theta(z^{(i)}, y^{\text{txt}}(P)) \odot \Sigma_{p\in P}W_p(I) \right\|_2,
\label{eq:attnloss}
\end{equation}
where $A_\theta$ denotes the response in cross attention layers in network $\theta$, namely $\text{Softmax}(QK^\top)$. The attention loss is designed to encourage the model to focus on the regions that are relevant to the photographic concepts specified in the text prompt $y_\text{txt}(P)$.

The training pipeline is shown in \cref{fig:pipeline_stage3}, where only the text embedding is optimized, based on the model weights after autoencoder pretraining and noise prediction network training. While the original entries are kept intact, the newly added pseudo words are supervised by
\begin{equation}
    \mathcal{L} = \mathcal{L}_{\text{WR}}+\lambda_{\text{AT}}\mathcal{L}_{\text{AT}},
\end{equation}
where the hyper-parameters are chosen as $\lambda_{\text{AT}}=0.01$. The random combination training strategy, along with $\mathcal{L}_\text{AT}$, helps disambiguate the intertwined photographic concepts.
Otherwise, the learned pseudo word embeddings would be affected by the appearance of other photographic concepts, thus leading to the incorrect control of photographic concepts.

\begin{figure*}[h]
    \centering
    \includegraphics[width=\linewidth]{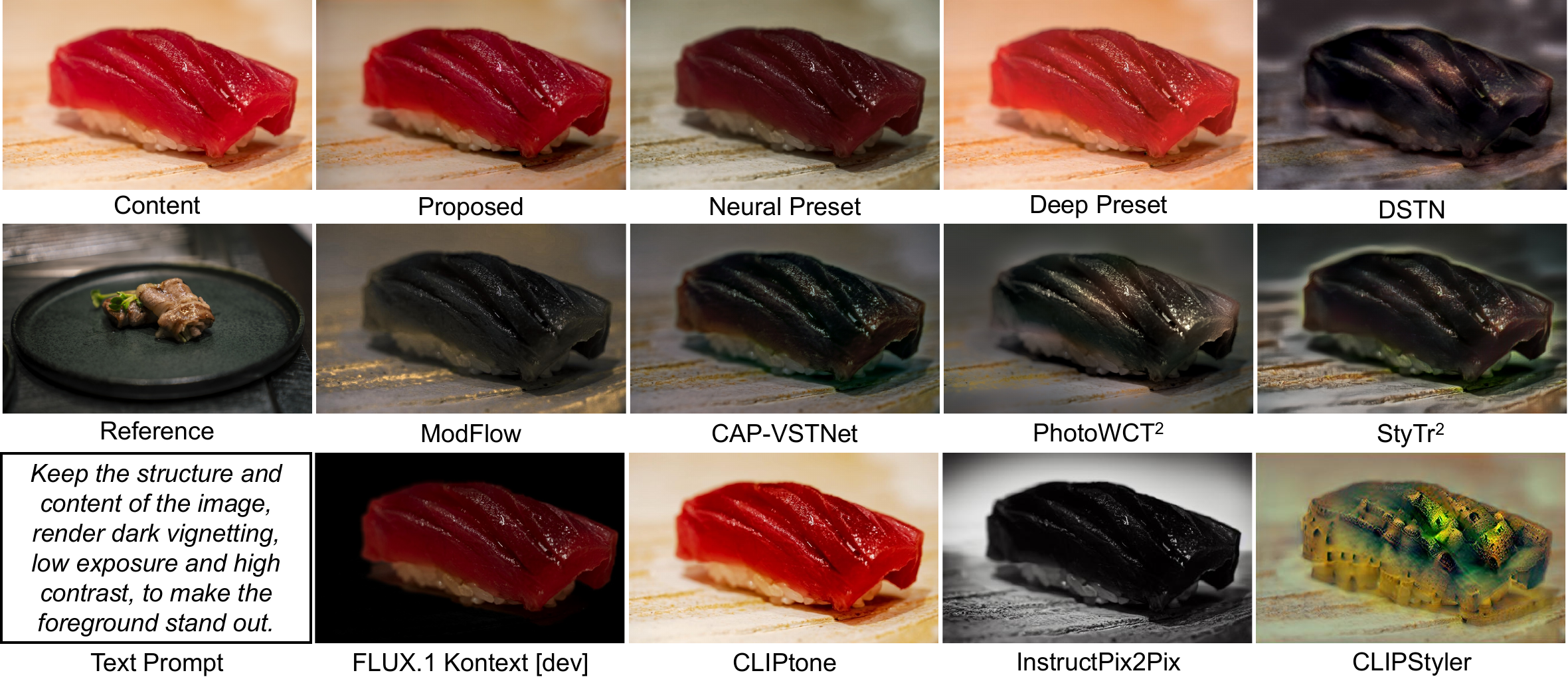}
    \caption{A qualitative comparison of the proposed method and the baselines methods, where the proposed method manages to transfer the vignetting and high contrast from the reference image, and keeps the sushi its original appealing color. In contrast, most image retouching methods mistakenly modify the color towards unsaturated dark color, except for Deep Preset~\cite{ho2021deep} that introduces minor changes to the color.}
    \label{fig:quali_sushi}
\end{figure*}
\begin{figure*}[h]
    \centering
    \includegraphics[width=\linewidth]{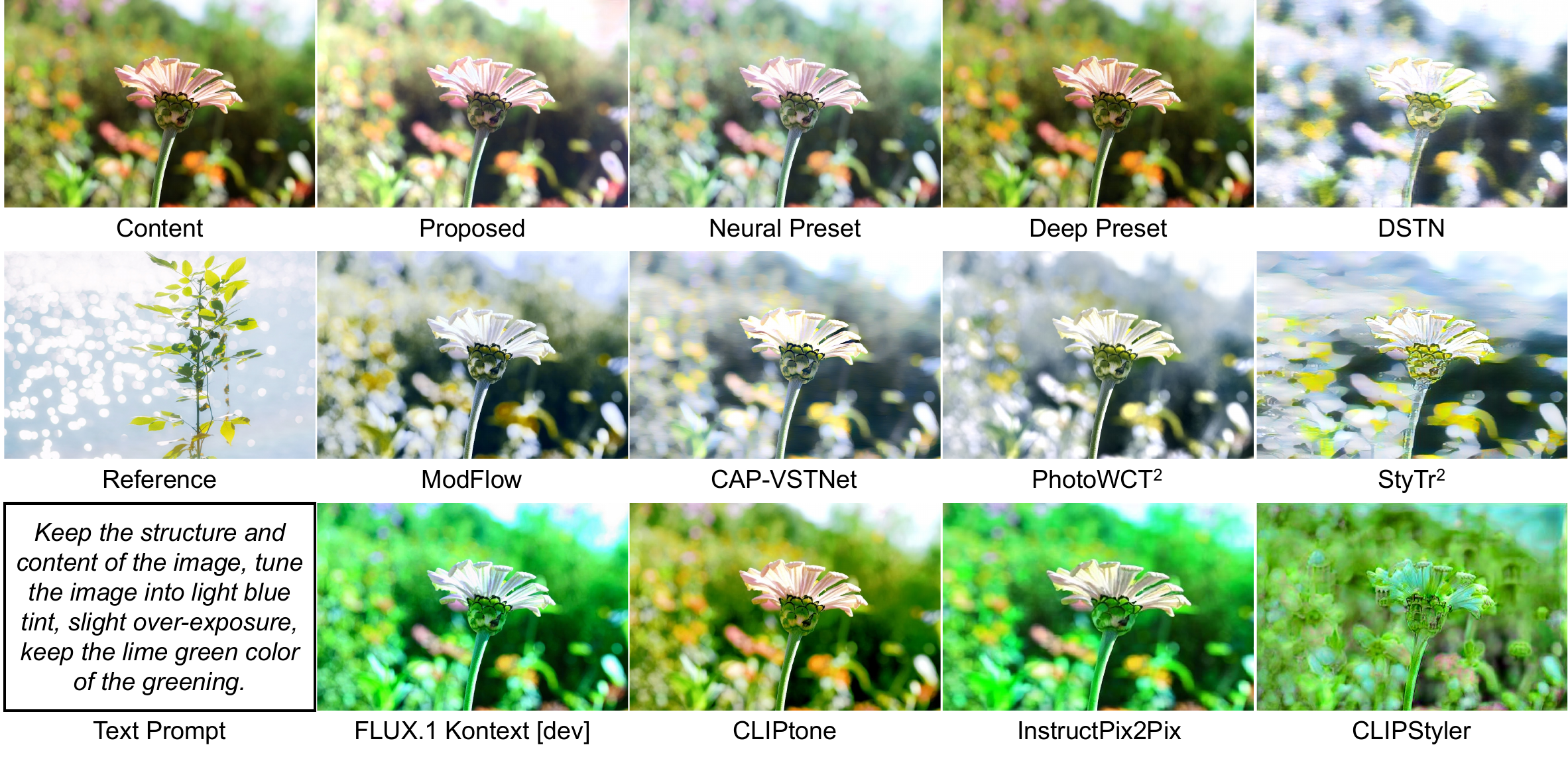}
    \caption{A qualitative comparison of the proposed method and the baselines methods. The proposed method learns to transfer the light blue tint from the reference image, while also aligning the lime green color of the plants. Most image retouching methods can only render the whole frame in the light blue color, but the colors from the content image is lost. Deep Preset~\cite{ho2021deep} introduces very minor changes, and Neural Preset~\cite{Ke2023neuralpreset}.}
    \label{fig:quali_spring_cxy}
\end{figure*}
\begin{figure*}[t]
    \centering
    \includegraphics[width=\linewidth]{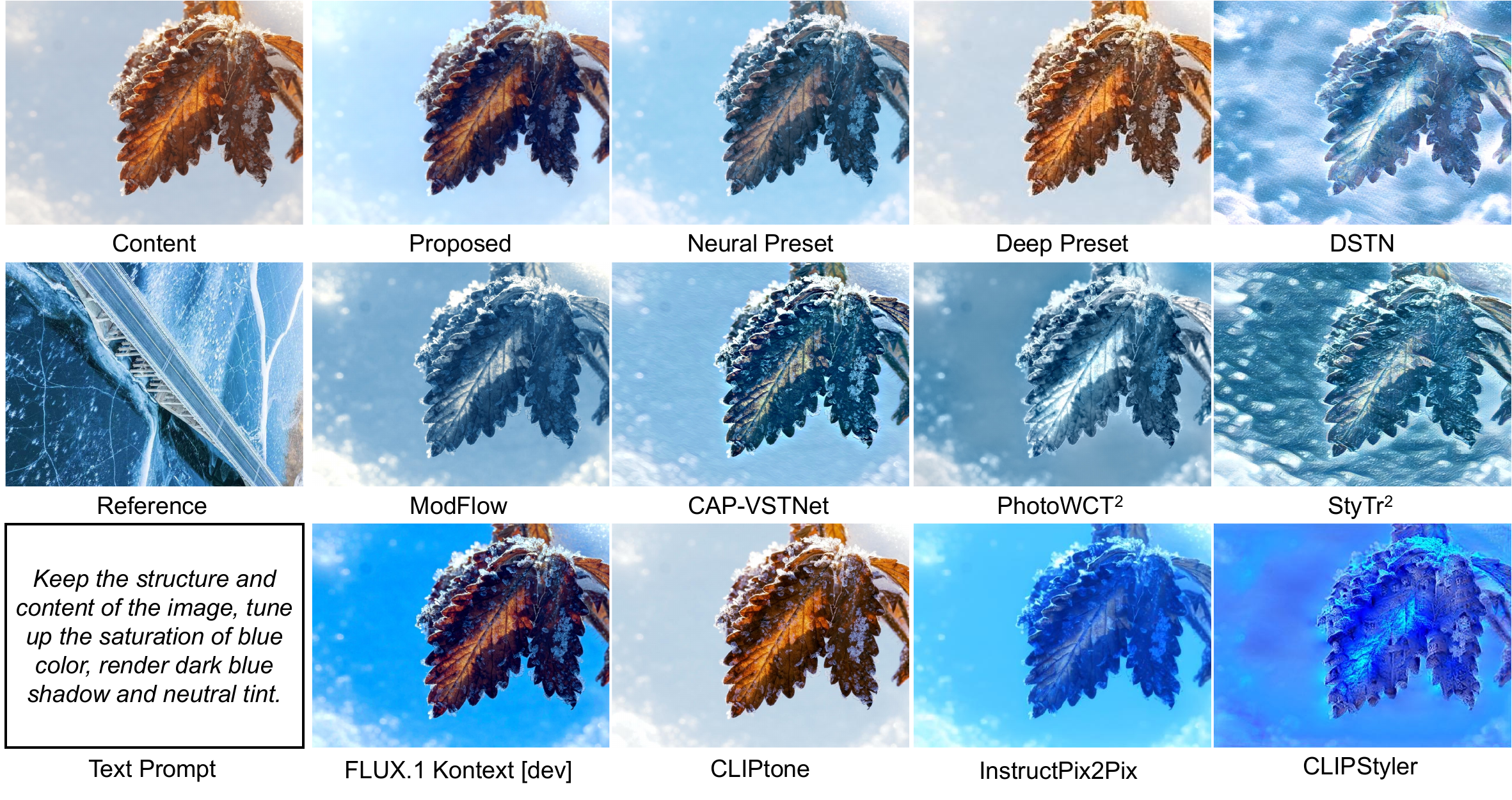}
    \caption{A qualitative comparison of the proposed method and the baselines methods. The leaves' brown color is only preserved by the proposed method and Deep Preset~\cite{ho2021deep}, while the latter fails to transfer the blue shadow embedded in the reference image. Other methods show an unrealistic bias towards blue by changing the color of the leaves completely.} %
    \label{fig:quali_ice_lfy}
\end{figure*}

\section{Experiments}\label{sec:experiments}
\subsection{Implementation Details}\label{sec:implementation}

\textit{(i)} \textbf{Datasets}: We gather a new dataset of photographic images from professional photographers as reference images, which will be released for academic use upon acceptance. 
In total, $20$ sets of photos have been collected, and the photos within each set are assumed consistent in style by the photographers themselves, with the number of photos ranging from $1$ to $5$. In addition, $3$ sets of photos come with an unedited version, which is the direct output of the default camera processing pipeline.

The content data is drawn from the COCO~\cite{lin2014microsoft} dataset for content images captured in diverse scenes, as well as the Unsplash~\cite{unsplash} dataset where we select photos with a similar genre to the reference images, for a more straightforward comparison of the transferred photographic style.

\textit{(ii)} \textbf{Training and inference}: All the experiments of PIF are conducted on 4 NVIDIA RTX A6000 GPUs. The residual one-step diffusion process is pretrained on Unsplash~\cite{unsplash} for 80k iterations, taking about 14 hours. The rank of LoRA is set to $8$. Note that the residal one-step diffusion only needs to be trained once, and is reused for photographic concept learning on any given reference images.

For photographic concept learning, The embeddings of the pseudo words to describe the photographic concepts are trained for 10k iterations, with an initial learning rate of $10^{-3}$ with cosine annealing, taking about 1 hour for each scene. The proposed method handles a variable number of reference images, even as few as a single image. The demonstrations within this paper involve no more than 5 reference images.

As we apply a one-step diffusion scheme, the inference time for a single image is comparable with previous CNN-based or transformer-based methods, as it only requires one denoising process per image.

\textit{(iii)} \textbf{Baselines}: Due to the great number of previous works related to reference-based image retouching, we select a few representative state-of-the-art open-source methods for comparison, including 
Deep Preset~\cite{ho2021deep}, PhotoWCT$^2$~\cite{chiu2022photowct2}, CAP-VSTNet~\cite{wen2023cap}, Neural Preset~\cite{Ke2023neuralpreset} and ModFlow~\cite{larchenko2025modflow}, all of which are specifically designed for photorealistic style transfer.  For a broader context, we also include DSTN~\cite{hong2021domain} and StyTr$^2$~\cite{StyTr2}, which are tailored for universal or artistic style transfer tasks.

Additionally, text-based image editing methods are considered for a more comprehensive evaluation, including FLUX.1 Kontext [dev]~\cite{labs2025flux1kontextflowmatching}, CLIPtone~\cite{lee2024cliptone},  CLIPStyler~\cite{kwon2022clipstyler} and InstructPix2pix~\cite{brooks2023instructpix2pix}. The descriptions of the reference images are given by the photographers themselves, and slightly simplified to work on the diffusion methods. Note that PIF does not use text prompts as input. 

For a fair comparison, we exclude methods that require fine-tuning on paired data with input and ground-truth images (\eg, MIT-Adobe FiveK dataset~\cite{fivek}), such as NeurOp~\cite{wang2022neural}, RSFNet~\cite{ouyang2023rsfnet}, DiffRetouch~\cite{duan2025diffretouch}, and SA-LUT~\cite{gong2025salutspatialadaptive4d}, because the input images do not contain the unedited version.
Also note that though we include all the baselines in qualitative comparisons, Neural Preset~\cite{Ke2023neuralpreset} is omitted in quantitative comparison, because it only provides access through the phone application, which makes it inaccessible to test at scale.

\subsection{Qualitative Results} 
We first show some cases where the content image shares a similar genre with the reference images. \cref{fig:quali_sushi,fig:quali_spring_cxy,fig:quali_ice_lfy} are organized in the same layout. The first columns are input conditions, and the others are generated results. The text prompts are given by the photographers themselves, and serve as the only condition for the methods in the last row. For the reference-based image editing methods in the first two rows, the reference image is the only condition.

In \cref{fig:quali_sushi}, the reference image shows dark vignetting and high contrast, with a strong focus on the wagyu beef in the plate. A similar photographic style is observed in the result of the proposed method. 
In comparison, most reference-based image retouching methods change the inherent appealing color of the sushi, while Deep Preset~\cite{ho2021deep} does not grasp the desired photographic style. For text-based image editing methods, FLUX.1 Kontext~\cite{labs2025flux1kontextflowmatching} and CLIPtone~\cite{lee2024cliptone} fail to understand ``vignetting'', and the former turns the background into darkness, while the latter keeps the vignetting intact. InstructPix2Pix~\cite{brooks2023instructpix2pix} is able to apply the dark vignetting, but the original color is diminished. CLIPStyler~\cite{kwon2022clipstyler} completely alters the image content.

In \cref{fig:quali_spring_cxy} where the slight over-exposure, low contrast, and light blue tint constitute the main photographic style, the propose method manages to learn that from the only one reference image. Among the reference-based image editing methods, Neural Preset~\cite{Ke2023neuralpreset} achieves the second best visual result, but it changes the greening in the background to a more blue color. For text-based image editing, CLIPtone~\cite{lee2024cliptone} turns the tint towards more green, and the others overshoots even more. The baselines cannot match the photographic concepts observed from the reference image.

In \cref{fig:quali_ice_lfy}, the reference image shows a frozen river, with a high saturation in blue, while the tint and highlight is neutral. The proposed method changes the color of the sky and the shadow of the leaves similar to the reference image, and keeps the original color of the brown leaves. In contrast, reference-based image retouching methods mostly wash the brown color away, while Deep Preset~\cite{ho2021deep} only slightly increases the exposure, failing to transfer the photographic style of the reference image.
Except for CLIPtone~\cite{lee2024cliptone} that seems to only increase the contrast, the results of other text-based image editing methods are biased towards more saturation than observed from the reference image, mostly because the text prompt does not specify the extent of the blue saturation.

\subsection{Quantitative Results}\label{sec:quantitative}
For quantitative measures, we test on 3 sets of paired data. By applying PIF to the unedited version of the test set, and comparing the generated images with the ground truths, we calculate the peak signal-to-noise ratio (PSNR), structural similarity index (SSIM)~\cite{ssimcite}, LPIPS~\cite{lpipscite} score, and the earth-mover distance~\cite{emdcite} (EMD) of the histogram to validate the accuracy of the proposed method.
PSNR measures the per-pixel accuracy, and SSIM~\cite{ssimcite} focuses on the local structural similarity over small windows. LPIPS~\cite{lpipscite} measures the perceptual distance with neural features of VGG~\cite{vggcite}. The EMD~\cite{emdcite} of histograms indicate the distance of lightness and color distribution.

Note that for the task of personalized image filter, the metrics regarding high-level features can only measure the preservation of original content, but not the photographic style transfer performance, as they rely more on the content and semantics of the image. As shown in \cref{tab:quanti}, the content image, without any modification, ranks the first in LPIPS~\cite{lpipscite} and second in SSIM~\cite{ssimcite}.

In comparison, the metrics of PSNR and EMD~\cite{emdcite} focus on how well the transfer actually is, by measuring the per-pixel difference and overall histogram distance of the prediction and the ground truth.
As listed in \cref{tab:quanti}, the proposed method significantly outperforms others on both PSNR and EMD~\cite{emdcite}, showing that the proposed method can learn the underlying photographic concepts and recover the color information with a higher accuracy. 

Judging from the SSIM~\cite{ssimcite} and LPIPS~\cite{lpipscite} scores, among the most content-preserving editing methods are Deep Preset~\cite{ho2021deep} and PhotoWCT$^2$~\cite{gupta2024photorealistic}. Both of these methods are equipped with the design to maintain pixel-wise fidelity, either with residual CNNs~\cite{ho2021deep} or wavelet transformations~\cite{chiu2022photowct2}, but their ability to transform photographic style is limited. Trained with text-image pairs, CLIPtone~\cite{lee2024cliptone} can adapt to novel scenes with natural language descriptions, but falls short in preserving the original content. This is because its inherent ambiguity of content and style.

These results, aligned with the visual results, show that reference-based image editing methods either tend to overfit the color, or cannot effectively learn the color adjustments.

\begin{table}[t]
    \centering
    \caption{Quantitative comparisons of the baselines and PIF (first 4 columns), and the average user study rating (last column). $\uparrow$ ($\downarrow$) indicates higher (lower) values are better. The best and second best results are marked in \first{red} and \second{blue}.}
    \label{tab:quanti}
    \setlength{\tabcolsep}{5pt}
    \begin{tabular}{lccccc}
\toprule
    Method & PSNR$\uparrow$ & SSIM$\uparrow$ & LPIPS$\downarrow$ & EMD$\downarrow$ & Rating$\uparrow$ \\
    \cmidrule(r){1-5}\cmidrule(l){6-6}
    PhotoWCT$^2$~\cite{chiu2022photowct2} & 18.61 & 0.8244 & 0.1574 & 0.0764 & 2.07 \\
     StyTr$^2$~\cite{StyTr2} & 18.81 & 0.6148 & 0.4844 & 0.0638 & 1.86 \\
     CAP-VSTNet~\cite{wen2023cap}& 23.22  & 0.8380 & 0.1801 & \second{0.0576} & 2.13 \\
     DSTN~\cite{hong2021domain} & 16.39 & 0.5525 & 0.4519 & 0.0795 & 1.42\\
     Deep Preset~\cite{ho2021deep} & 22.58 & 0.8535 &  0.2052 &  0.0750 & 3.09 \\
     ModFlow~\cite{larchenko2025modflow} & 21.39 & 0.8179 & 0.2311 & 0.0770 & 2.91 \\
     InstructPix2Pix~\cite{brooks2023instructpix2pix} & 21.60 & 0.6408 & 0.3543 & 0.0832& 2.11 \\
     CLIPStyler~\cite{kwon2022clipstyler} & 16.64 & 0.3874 & 0.5957 & 0.1393 & 1.55 \\
     CLIPtone~\cite{lee2024cliptone} & \second{26.92} &	0.7331 & 0.3090  &	0.0851 & 3.91\\
    Flux.1 Kontext~\cite{labs2025flux1kontextflowmatching} & 22.65 & 0.7872 & 0.1463 & 0.0750 & \second{4.30} \\
     Proposed & \first{28.79} & \first{0.9079} & \second{0.1042} & \first{0.0374} & \first{4.53} \\
     \midrule
     Content & 22.48 & \second{0.8911} & \first{0.1019} & 0.0754 & N/A \\
 \bottomrule
    \end{tabular}
\end{table}

\subsection{User Study}\label{sec:user_study}

To judge whether the generated images share the same photographic concepts with the reference images is a tough task, as the photographic style is still not defined properly. Therefore, we conduct a user study to evaluate the effectiveness of the proposed method. The user study consists of $160$ rounds, with a total of $30$ amateurs and $15$ professional photographers. They are asked to give a rating out of 5 about the quality of transferring the photographic concept while preserving the contents. 

\begin{figure}
    \centering
    \includegraphics[width=\linewidth]{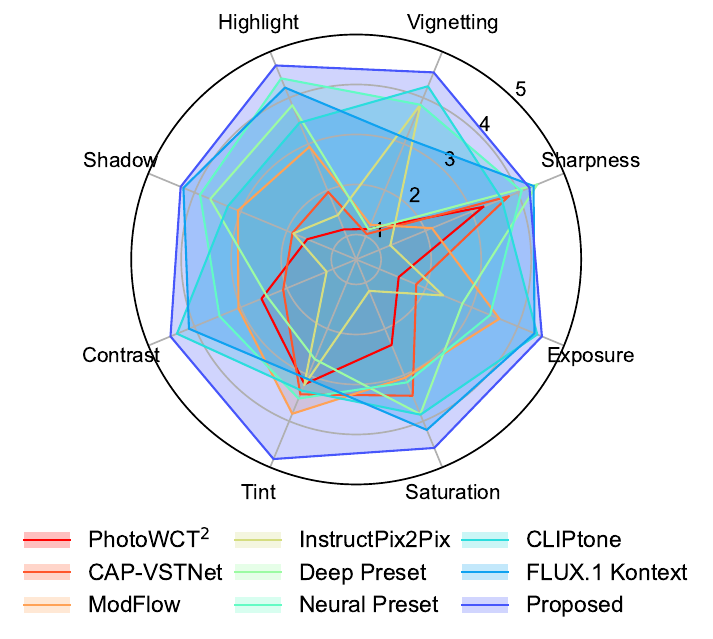}
    \caption{The average rating of each photographic concept, given by the professional group in the user study.}
    \label{fig:popularity}
\end{figure}

The average rating for each method is listed in the last column of \cref{tab:quanti}, and it shows that PIF is the most preferred method in learning and transferring photographic concepts.
We also show the detailed rating of the professional photographers across all photographic concepts in \cref{fig:popularity}. The photographic concepts of tint, vignetting, saturation, and highlight of PIF are especially recognized.

\subsection{Ablation Study}\label{sec:ablation}
\begin{figure*}[h]
    \centering 
    \includegraphics[width=1.0\linewidth]{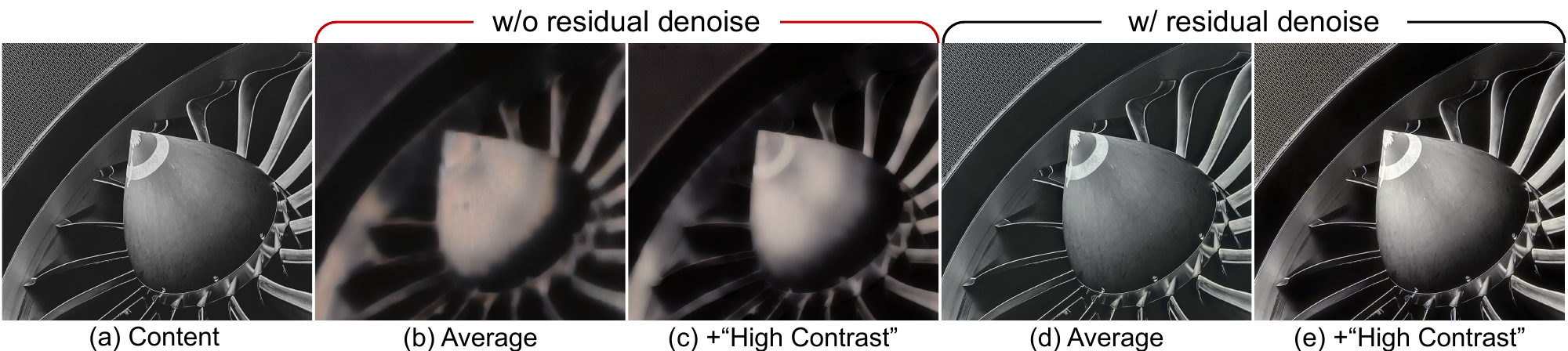}
    \caption{The ablation study of the residual denoising paradigm proposed in \cref{eq:res_denoise}. Given (a) content image and the text prompts $y_{\text{txt}}$ as a combination of ``average'' photographic concepts, the residual denoising is able to synthesize (d) the average appearance, without which (b) the average appearance lacks in detail severely. In (c) and (e), only the concept of contrast is modified to ``high'', and the residual denoising also outperforms in instruction following and content preservation.}
    \label{fig:ablation-res-denoise}
\end{figure*}

First, we validate the proposed residual denoising scheme in \cref{eq:res_denoise}. The comparison in \cref{fig:ablation-res-denoise} shows that the proposed residual denoising scheme excels the vanilla denoising in structure preserving. Though the vanilla denoising is also able to adjust the contrast slightly after trained on the same objective, the issue of detail loss cannot be addressed.

\begin{figure}
    \centering 
    \includegraphics[width=1.0\linewidth]{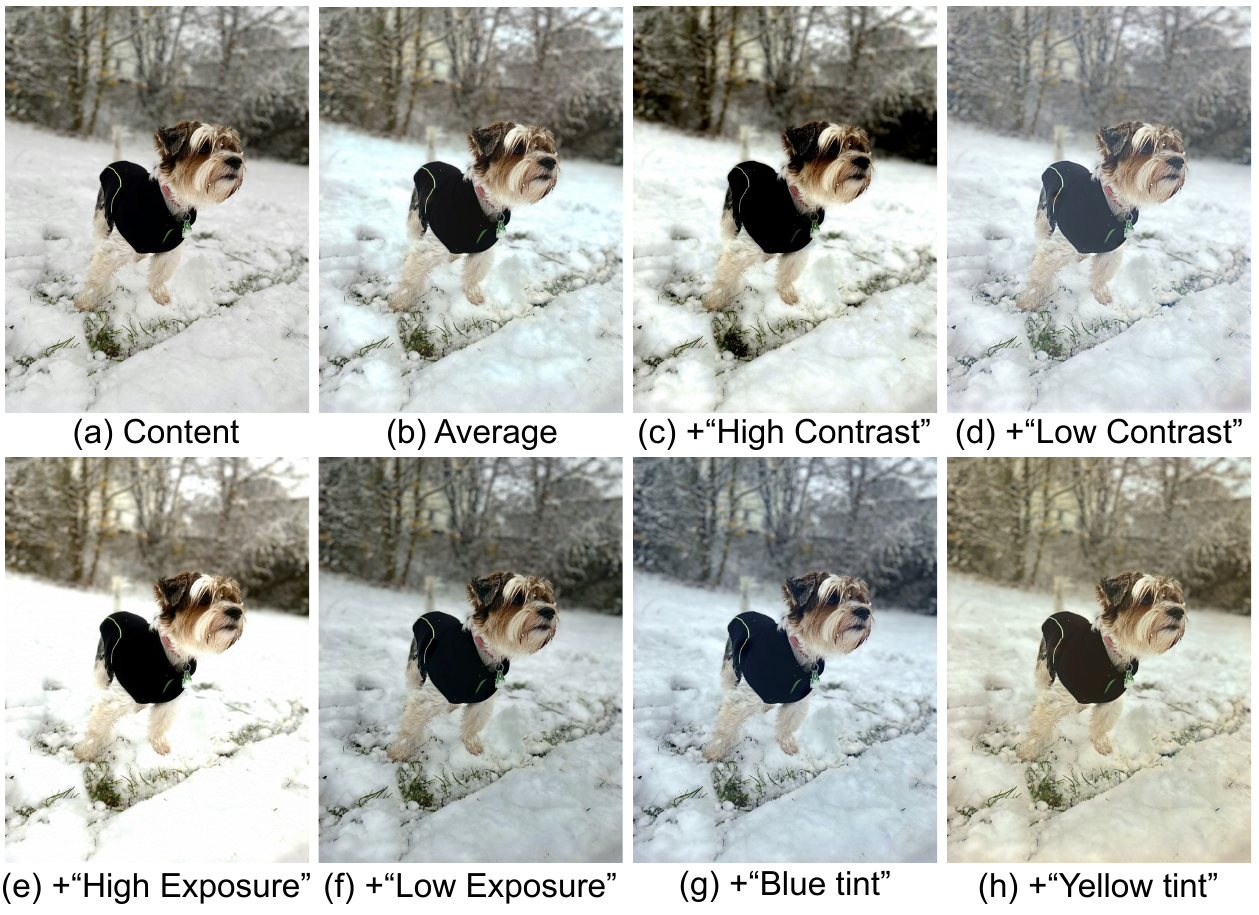}
    \caption{The residual one-step diffusion results to edit (a) content image. The text conditions $y_\text{txt}$ are (b) the combination of average photographic concepts, and (c)-(h) the combination with only one of the photographic concepts adjusted based on the average prompts. This demonstrates that the photographic style can be controlled by the proposed residual one-step diffusion, and the detail change is minimal.}
    \label{fig:prompt-visual}
\end{figure}

We then verify the ability to control the photographic style with text prompts, as illustrated in \cref{fig:prompt-visual}. Given \cref{fig:prompt-visual}(a) as the content image, the residual one-step diffusion trained in \cref{sec:one_step} is able to modify the photographic style only, and leave the content of the image unchanged. Starting from the average appearance, the trained residual one-step diffusion model is able to adjust the photographic concepts (\eg, contrast, exposure, and tint) with respect to the text prompts.

\begin{figure*}
    \centering
    \includegraphics[width=\linewidth]{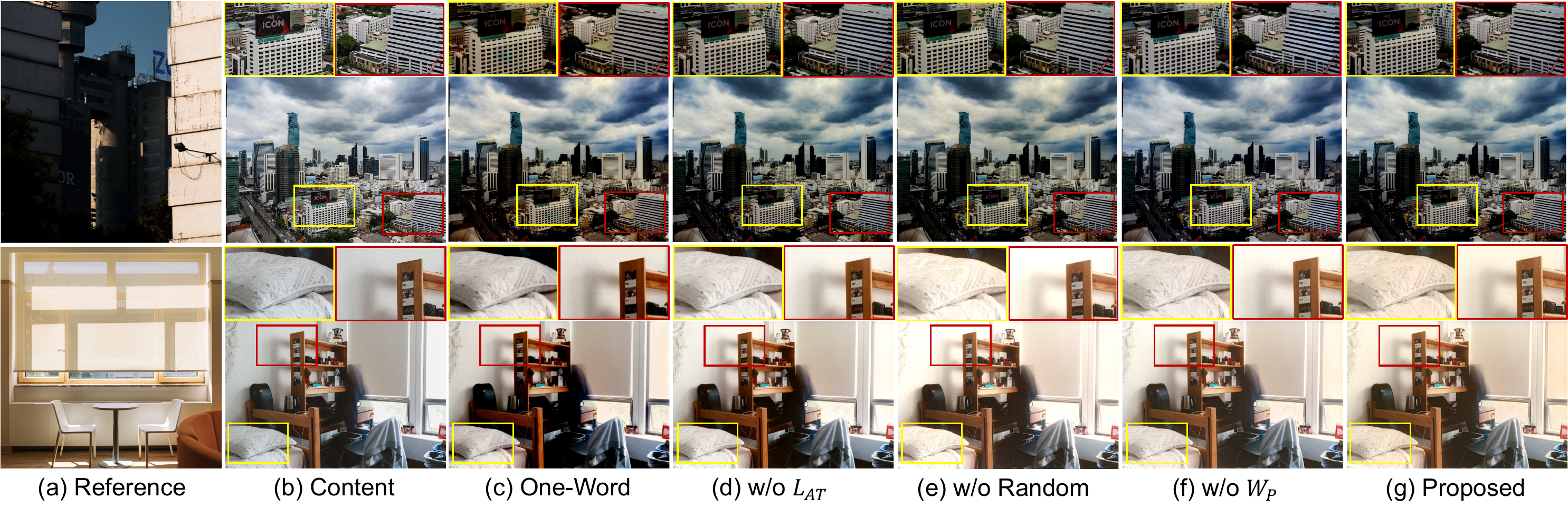}
    \caption{The ablation study of PIF. For the two (a) reference scene with given (b) content images, we compare (g) the complete PIF with (c) only optimizing one pseudo-word to represent the photographic style, (d) removing the weight functions, (e) removing the random combination training strategy, (f) removing the attention loss defined in \cref{eq:attnloss}. PIF achieves the best visual results significantly, verifying the effectiveness of the modules.}
    \label{fig:ablation}
\end{figure*}

We then verify the designs involved when optimizing photographic style in the cases shown in \cref{fig:ablation}. The first case calls for the style of a cold tint and shadow, high contrast, and unsaturated orange highlight, while the content image shows a cityscape on a cloudy day; and the second one conveys a warm tint, with low saturation, unsaturated orange highlight, and softened light shadow, with a content image of a white-balanced indoor room.
Since the genres of the reference images and content images are close to each other, all of the photographic style is expected to be transferred.

The results of only optimizing one pseudo-word to describe the photographic style is depicted in \cref{fig:ablation}(c). The ablation setting leads to a dark blue spot artifact in the first case (on the building, lower middle), as well as adding the contrast by mistake in the second case, suggesting the deficiency of representing photographic style with only one word.

We then discuss the influence of the attention loss $\mathcal{L}_\text{AT}$ in \cref{fig:ablation}(d). With the attention loss removed, the binding between photographic concepts and pseudo words becomes less stable. The results show a similar dark spot in the first case, and contrast in the second example is wrongly increased.

In \cref{fig:ablation}(e), we validate the random combination training strategy by applying all concepts in the text prompts throughout training. The results show a yellow tint in the first case, and a over-exposed highlight in the second, deviating from the reference images. This indicates that the embeddings for the photographic concepts are not correctly learned due to the lack of random combination training strategy.

In \cref{fig:ablation}(f), we remove the weight functions $W_p$, and the supervision of photographic concept learning degenerates to the naive reconstruction loss. The model is able to learn some of the photographic concepts, but tends to mix different concepts together. For example, 
and the wall in the second example is turned to pink due to the error.

\subsection{Discussion}\label{sec:discussion}
\begin{figure*}
    \centering
    \includegraphics[width=\linewidth]{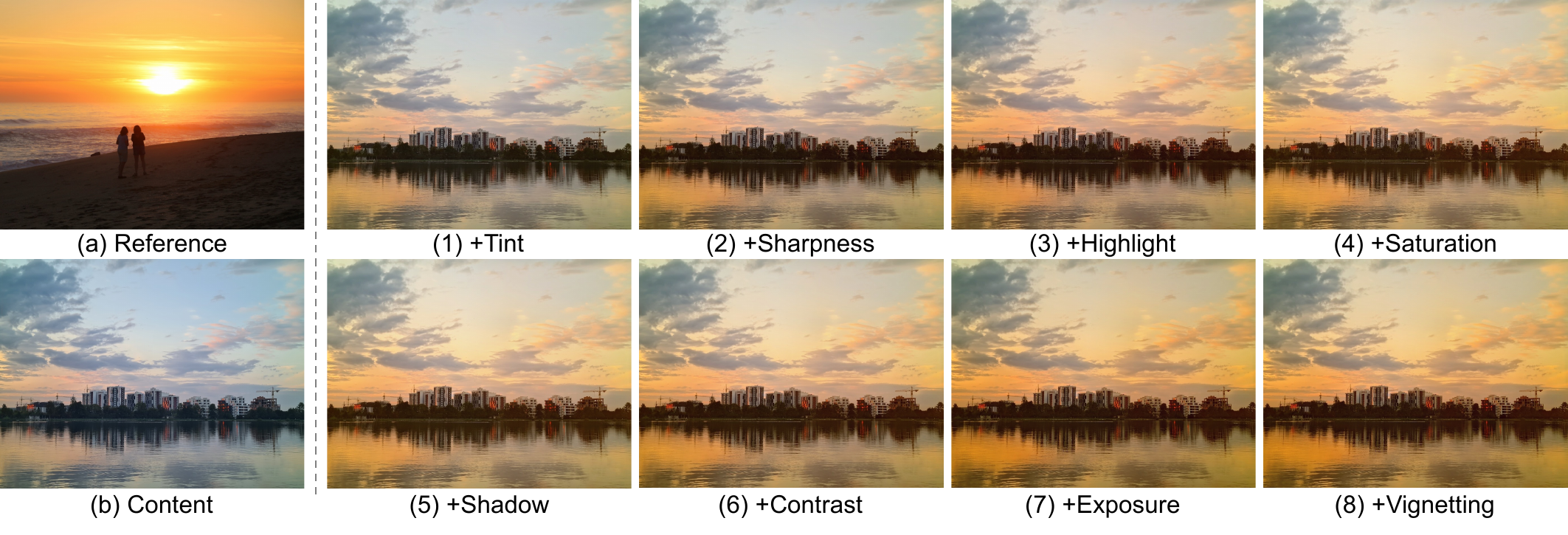}
    \caption{With the concepts optimized from (a) reference image, PIF renders them onto (b) content image incrementally. From (1) to (8), in each photo, the optimized concept in the caption substitutes the average concept presented in the prompts. As the photos gradually approach the reference photographic style in all the aspects, the white-box nature of PIF is demonstrated.}
    \label{fig:partial}
\end{figure*}
For the mission of image filters, white-box models are more preferred than black-box ones, as they enable the users to control each photographic concepts interactively. The implicit separation of various photographic concepts has endowed PIF with such ability.

As shown in \cref{fig:partial}(a), the reference image has a yellow highlight and warm tint, with the shadow part is affected by the sun glow, and thus colored in orange. It also has a slightly dark vignetting, reduced sharpness (observed from the sun glow), and low exposure and average contrast (judging from the visible details in the shadow). With most of these concepts are absent from the content images in \cref{fig:partial}(b), we simulate the process of progressively adding more concepts. 

Starting from the content image, with the photographic embedding added incrementally, the results show the corresponding photographic concept. In especial, PIF is able to significantly shift the tint, highlight, and exposure to the similar level to the reference image. Its adjustment becomes subtle when the photographic concept is already similar to the reference image, such as the sharpness and saturation.

\begin{figure*}[t]
    \includegraphics[width=\linewidth]{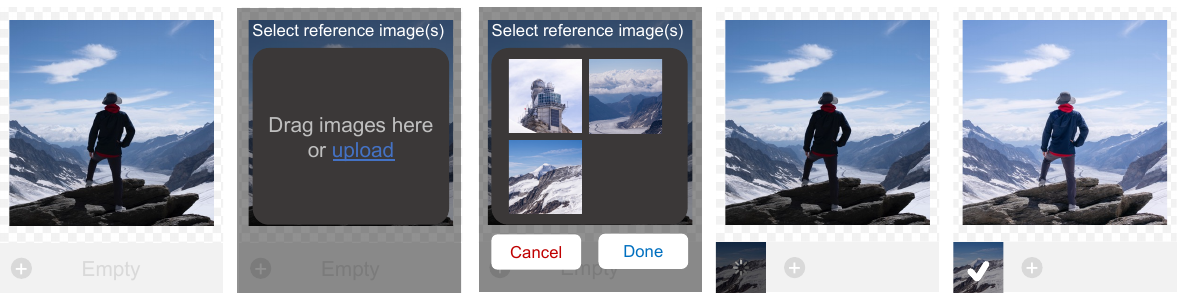}
    \caption{The graphic user interface (GUI) design of the PIF application.} \label{fig:GUI}
\end{figure*}

In addition, we built an application with a graphic user interface, and called for test users to personalize their own photographic style with uploaded photos of their own, or their preferred photographers. The process of using the application is shown in \cref{fig:GUI}. The photographic style learning is performed in a cloud server, as it requires computation-intensive optimization. After the optimization is completed, the word embedding of the pseudo words are transmitted back to the device, and the application can render the learned photographic style to any content image.

\begin{figure*}
    \centering
    \includegraphics[width=\linewidth]{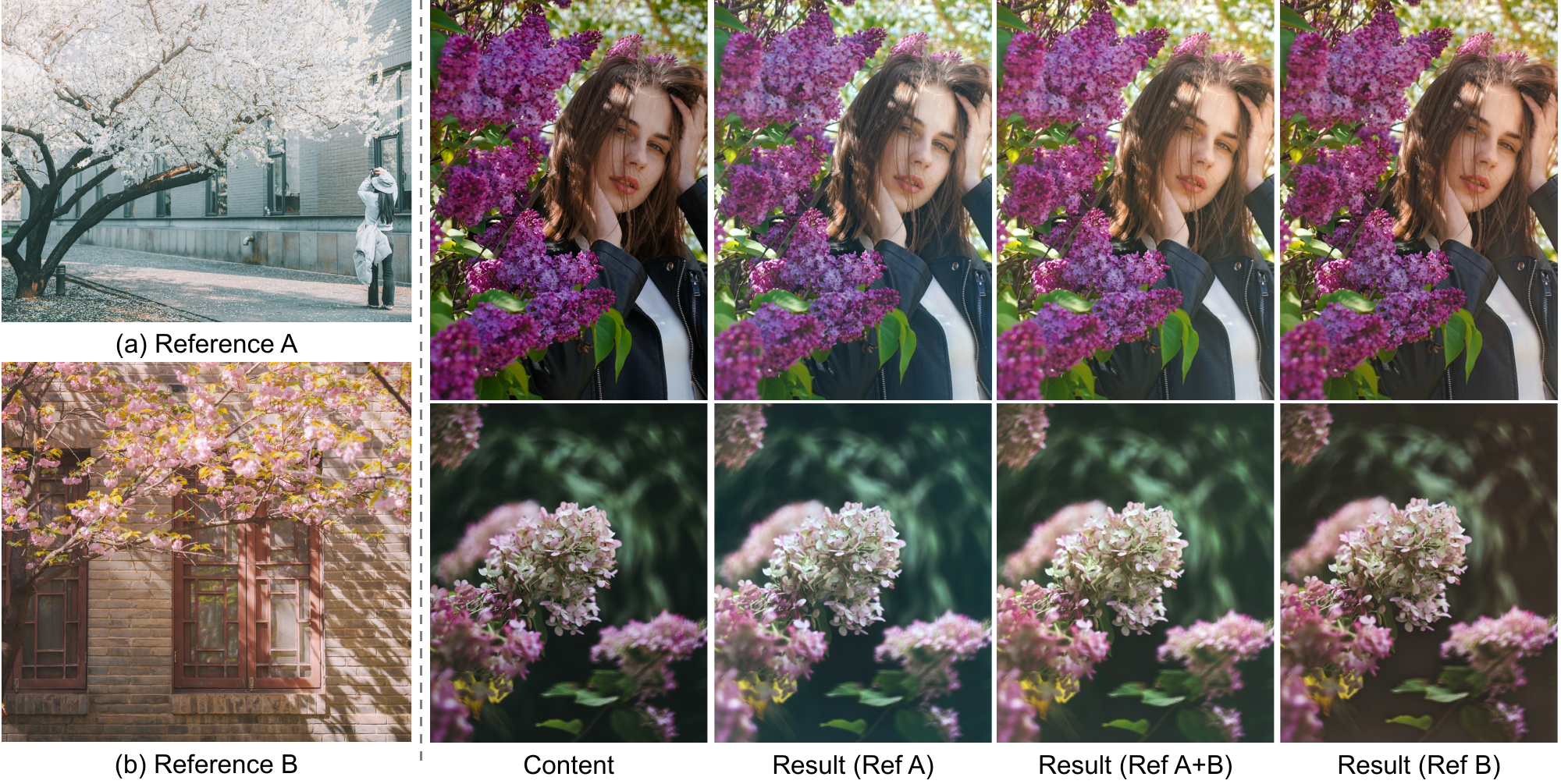}
    \caption{The results of learning reference images with conflicting photographic concepts (Ref A+B), in comparison with optimizing with either one of the reference images (Ref A and Ref B).}
    \label{fig:demo-conflict}
\end{figure*}

Last but not least, the reference images may contain conflicting photographic concepts, apart from the shared concepts deemed as the photographic style. PIF learns the photographic concepts in a similar way to summing up vectors. As the optimization only involves text embedding of pseudo words, and word embeddings exhibit a linear behavior~\cite{levy2014neural,mikolov2013linguistic}, conflicting concepts cancel each other out during optimization, and the embeddings of those modifiers are close to that of ``average'', as shown in \cref{fig:demo-conflict}. The image of ``reference A'' (Ref A) shows a blue tint, light shadow, low contrast, and fuzzy highlight. ``Reference B'' (Ref B) shares the photographic style in contrast, highlight, sharpness and shadow, but differs in the tint, by leaning towards orange.
PIF tunes up the dark shadow in the two content images, decreases the contrast, and adds fuzzy brightness to the highlight part, namely the light blobs on the person's hair and the flower blossoms. Comparing (Ref A) and (Ref B), the result of (Ref A+B) presents an interpolated effect between (Ref A) and (Ref B).

\section{Conclusion}\label{sec:conclusion}
We formulate the task of transferring photographic style and a novel Personalized Image Filter (PIF). PIF can master the professional photographic style by learning the photographic concepts from a set of reference images with a random combination training strategy. Then it can transfer the learned photographic style to any content images while preserving the contents of the latter, with the proposed residual one-step diffusion. 
The defined photographic concepts are also flexible for improvement. Acting as a white-box model, PIF is expected to benefit users and the photography community. 

\textbf{Limitations}: Currently, PIF is unaware of object-level photographic concepts, but since similar masks can be made by object segmentation methods (\eg, SAM~\cite{kirillov2023segment}), the proposed pipeline can still work. PIF still changes some details due to the VAE compression, which can be solved by using a backbone with less compression in latent space~\cite{sd3,flux}.

\bibliographystyle{splncs04}
\bibliography{egbib}

\begin{IEEEbiography}[{\includegraphics[width=1in,height=1.25in,clip,keepaspectratio]{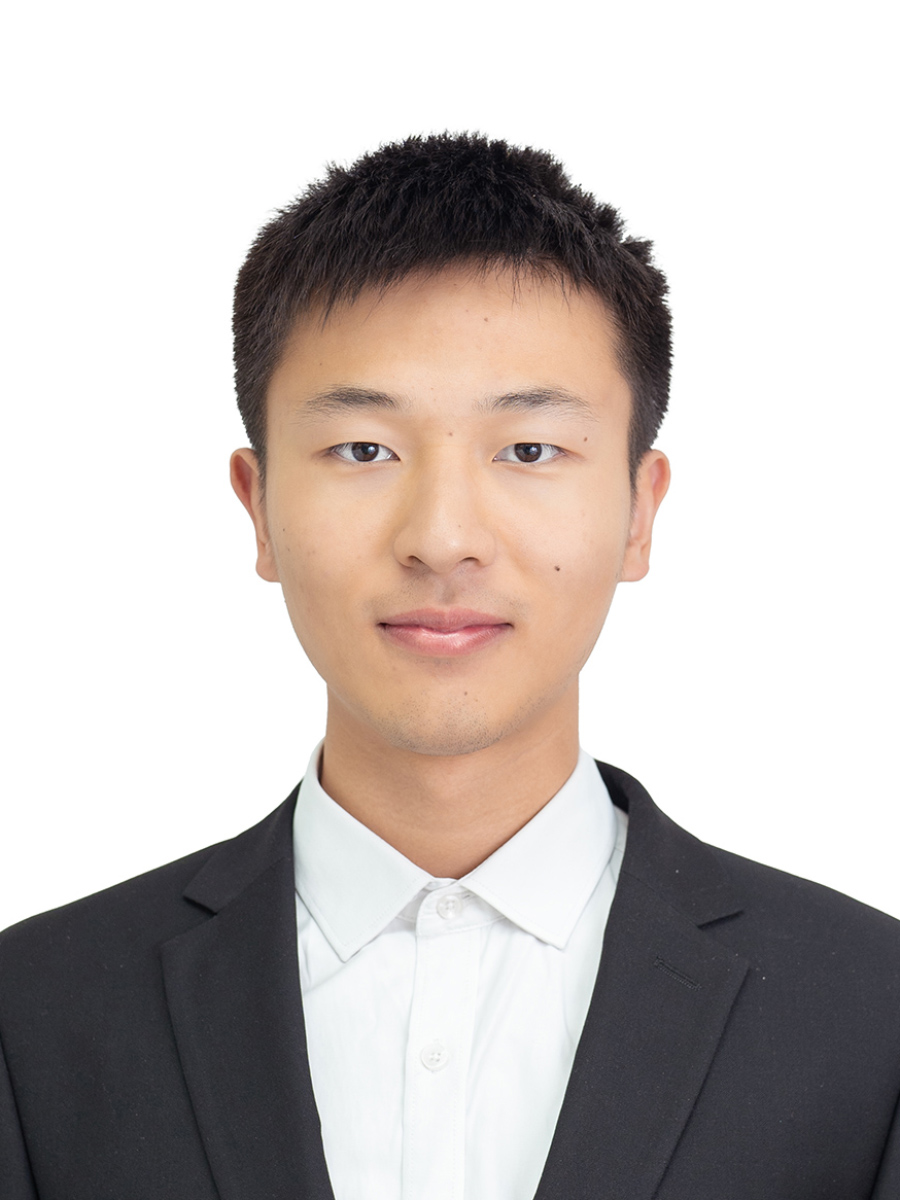}}]{Chengxuan Zhu}
is a Ph.D. student at the school of Intelligence Science and Technology, Peking University. He received the B.S. degree from Peking University in 2019. His research interest lies in combining computational photography with AIGC.
\end{IEEEbiography}

\begin{IEEEbiography}[{\includegraphics[width=1in,height=1.25in,clip,keepaspectratio]{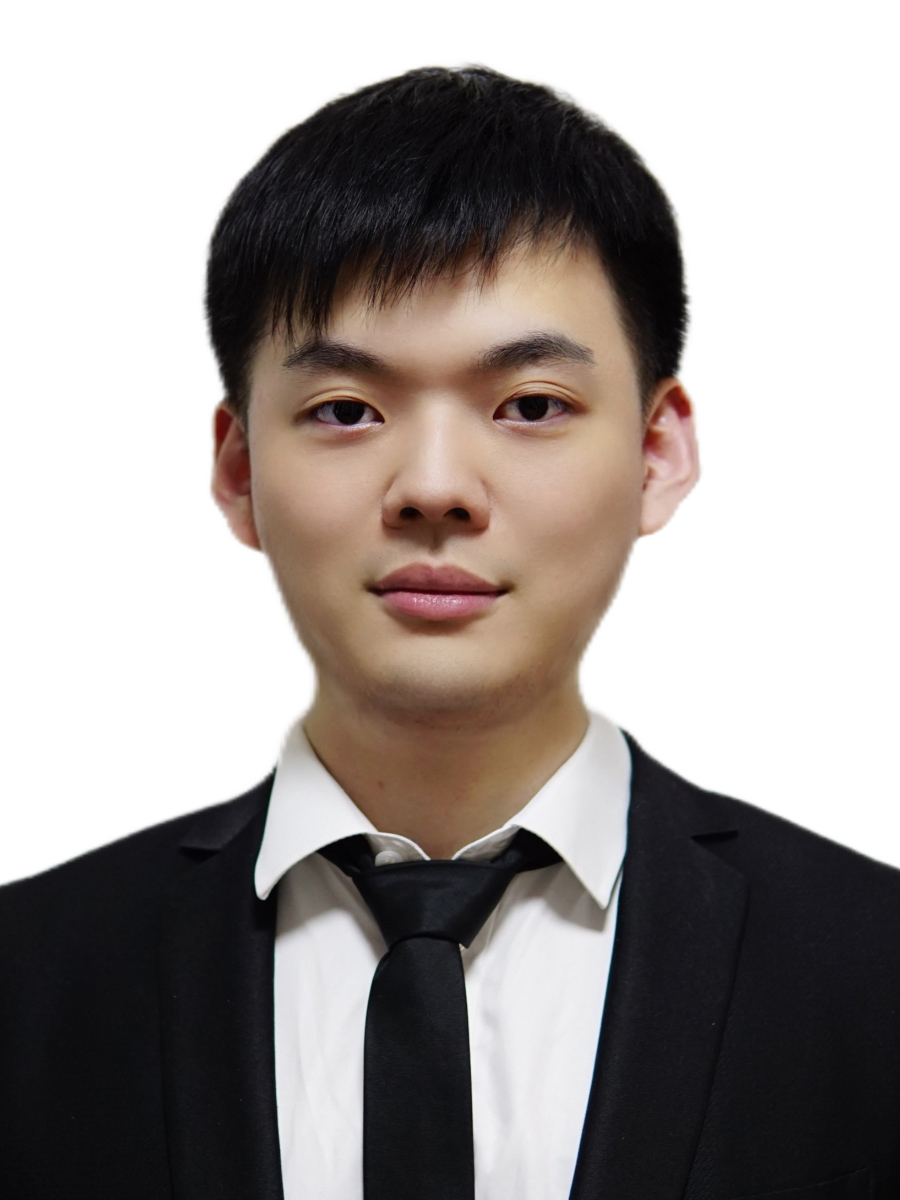}}]{Shuchen Weng}
is a research scientist at Beijing Academy of Artificial Intelligence. He received the Ph.D. degree from Peking University in 2024. His research interests include cross-modality (mainly language-based) content creation and manipulation.
\end{IEEEbiography}

\begin{IEEEbiography}
    [{\includegraphics[width=1in,height=1.25in,clip,keepaspectratio]{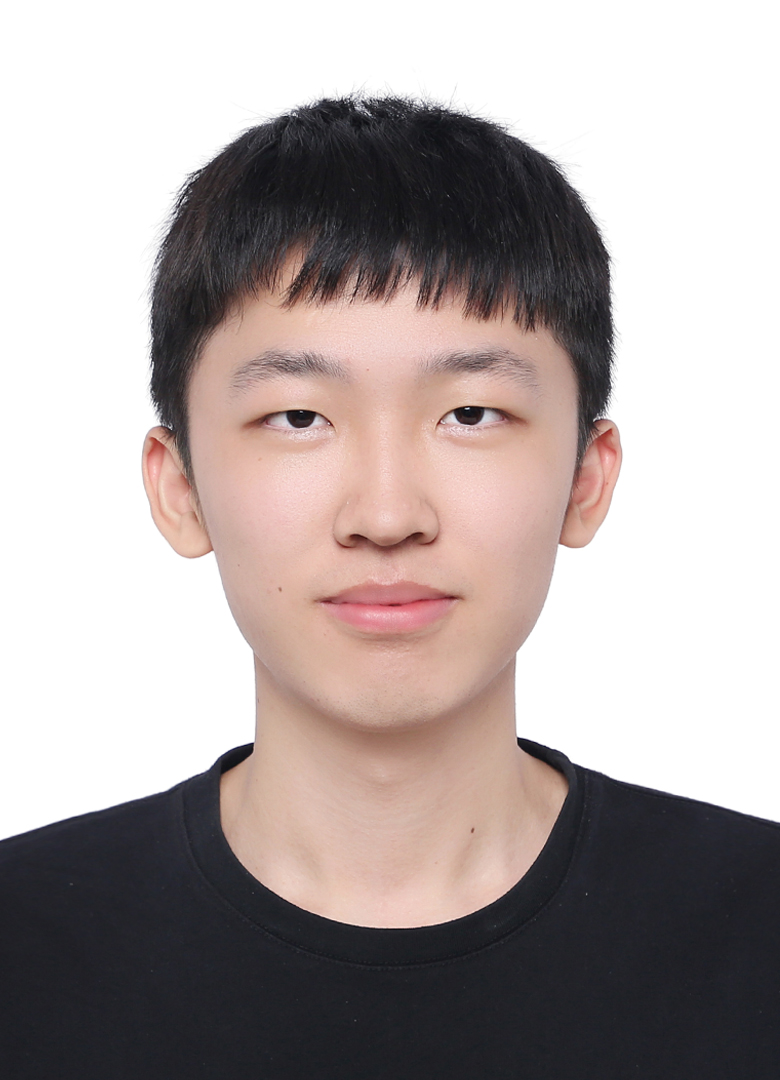}}]{Jiacong Fang} 
    is a senior undergraduate student at Yuanpei College, Peking University. 
    His current research interests include computational photography and the application of AIGC in controllable image editing.
\end{IEEEbiography}

\begin{IEEEbiography}
[{\includegraphics[width=1in,height=1.25in,clip,keepaspectratio]{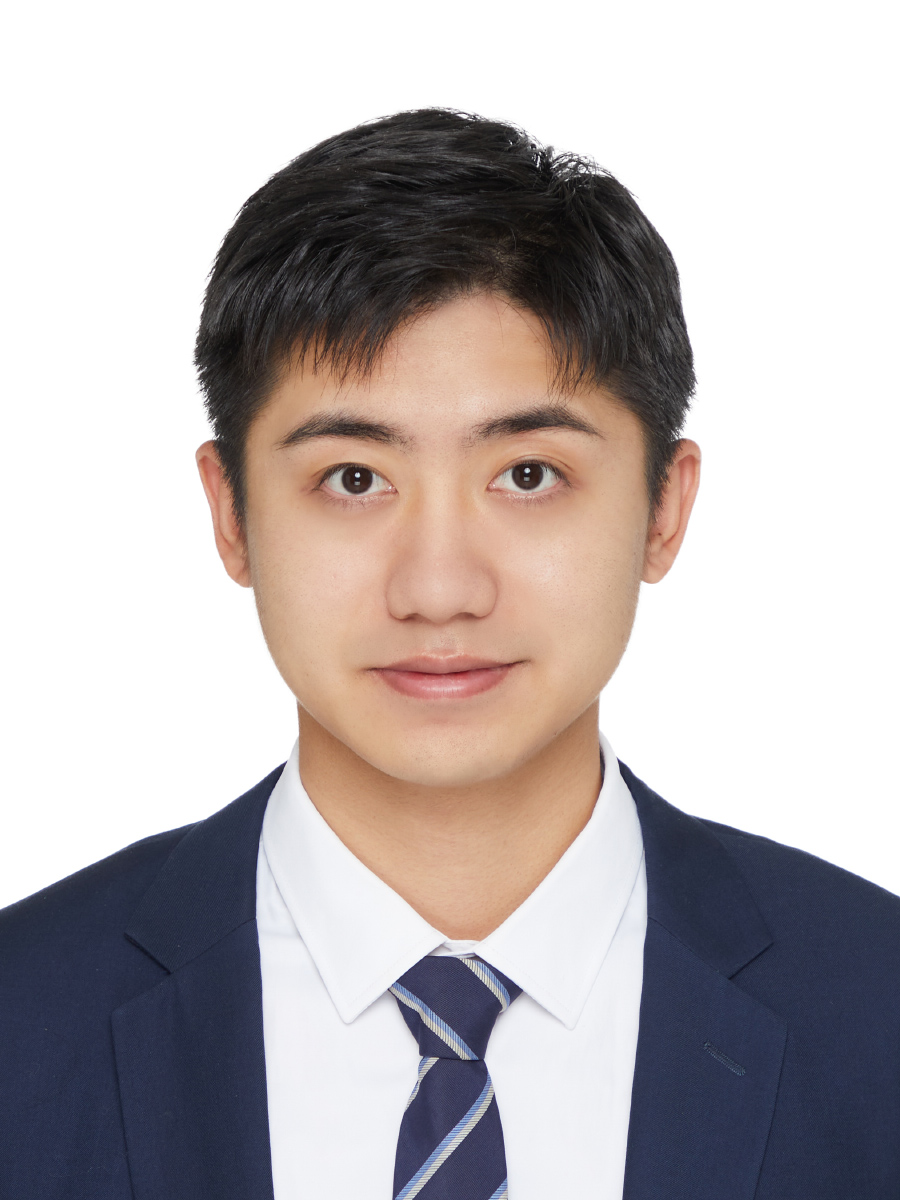}}]{Peixuan Zhang}
is a Ph.D. student at the School of Artificial Intelligence, Beijing University
of Posts and Telecommunications. His current research interests include include affective computing and cross-modality (mainly language-based) content creation.
\end{IEEEbiography}

\begin{IEEEbiography}[{\includegraphics[width=1in,height=1.25in,clip,keepaspectratio]{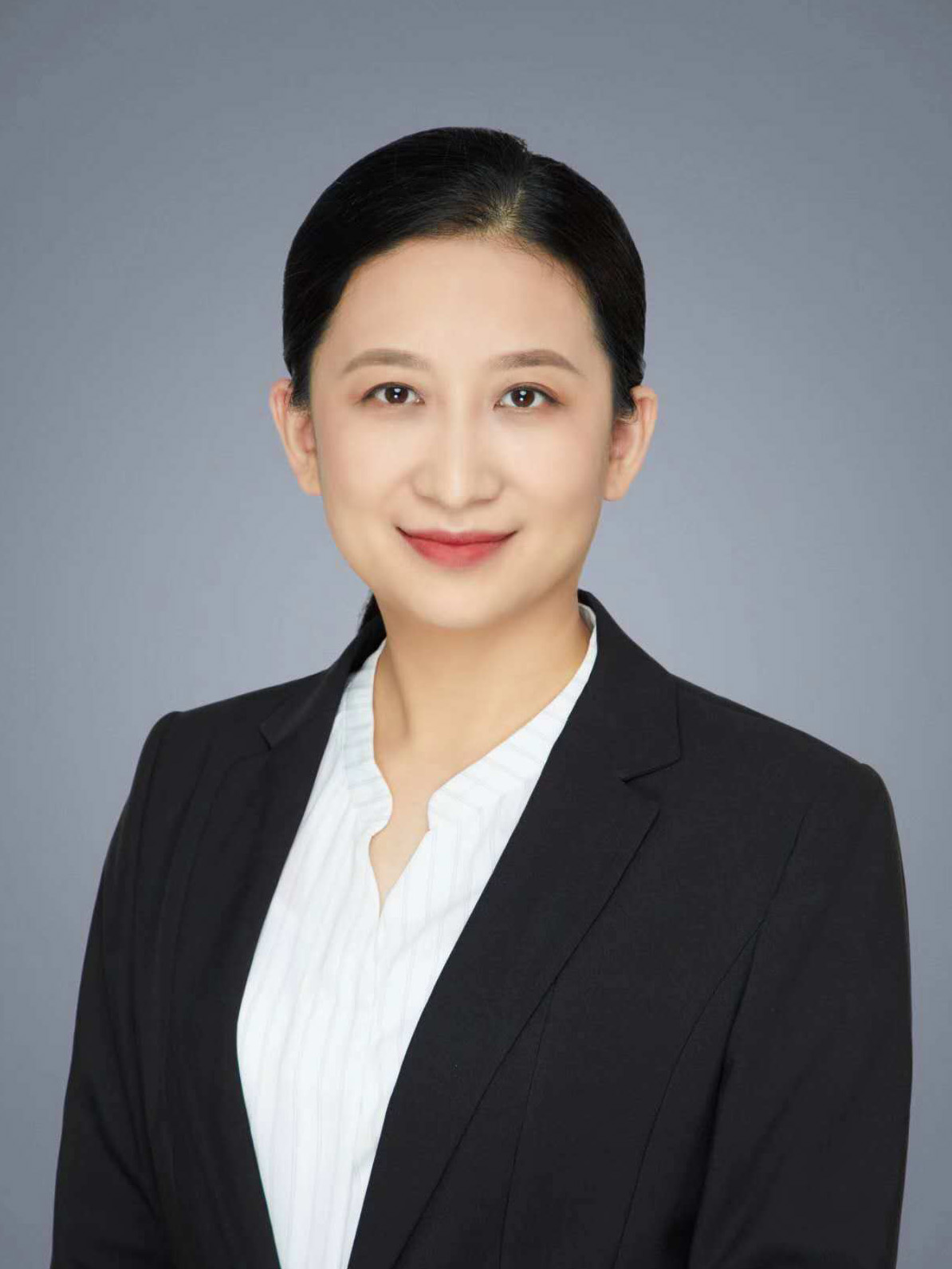}}]{Si Li}
received the Ph.D. degree from the Beijing University of Posts and Telecommunications in 2012. She is currently an associate Professor with the School of Artificial Intelligence, Beijing University of Posts and Telecommunications. Her current research interests include multimodal artificial intelligence and machine learning.
\end{IEEEbiography}

\begin{IEEEbiography}
[{\includegraphics[width=1in,height=1.25in,clip,keepaspectratio]{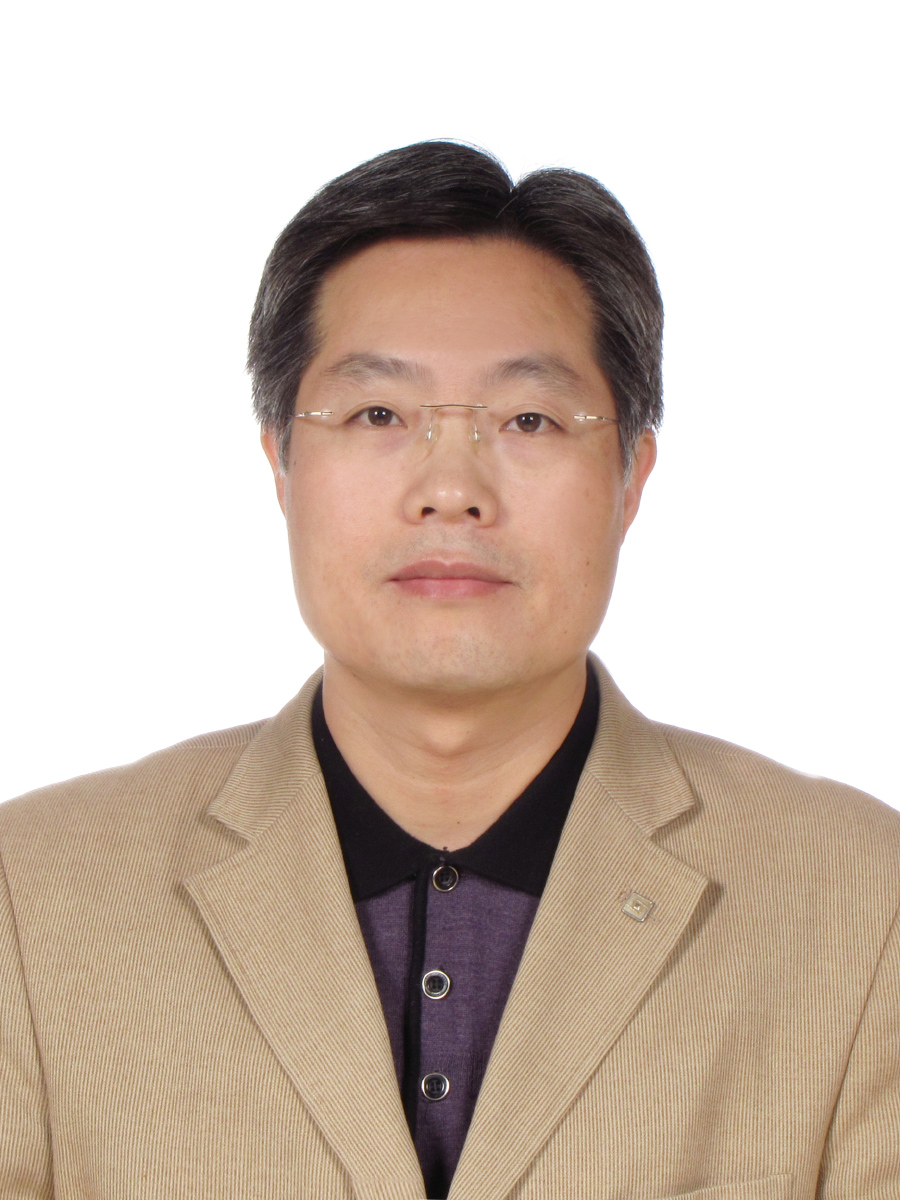}}]
{Chao Xu} (Member, IEEE) received the B.E. degree from Tsinghua University, Beijing, China, in 1988, the M.S. degree from the University of Science and Technology of China, Hefei, China, in 1991, and the Ph.D. degree from the Institute of Electronics, Chinese Academy of Sciences, Beijing, in 1997. From 1991 to 1994, he was an Assistant Professor with the University of Science and Technology of China. Since 1997, he has been with the School of Electronics Engineering and Computer Science (EECS), Peking University, Beijing, where he is currently a Professor. His research interests include image and video coding, processing, and understanding. He has authored or coauthored more than 80 publications and 5 patents in these fields.
\end{IEEEbiography}

\begin{IEEEbiography}
[{\includegraphics[width=1in,height=1.25in,clip,keepaspectratio]{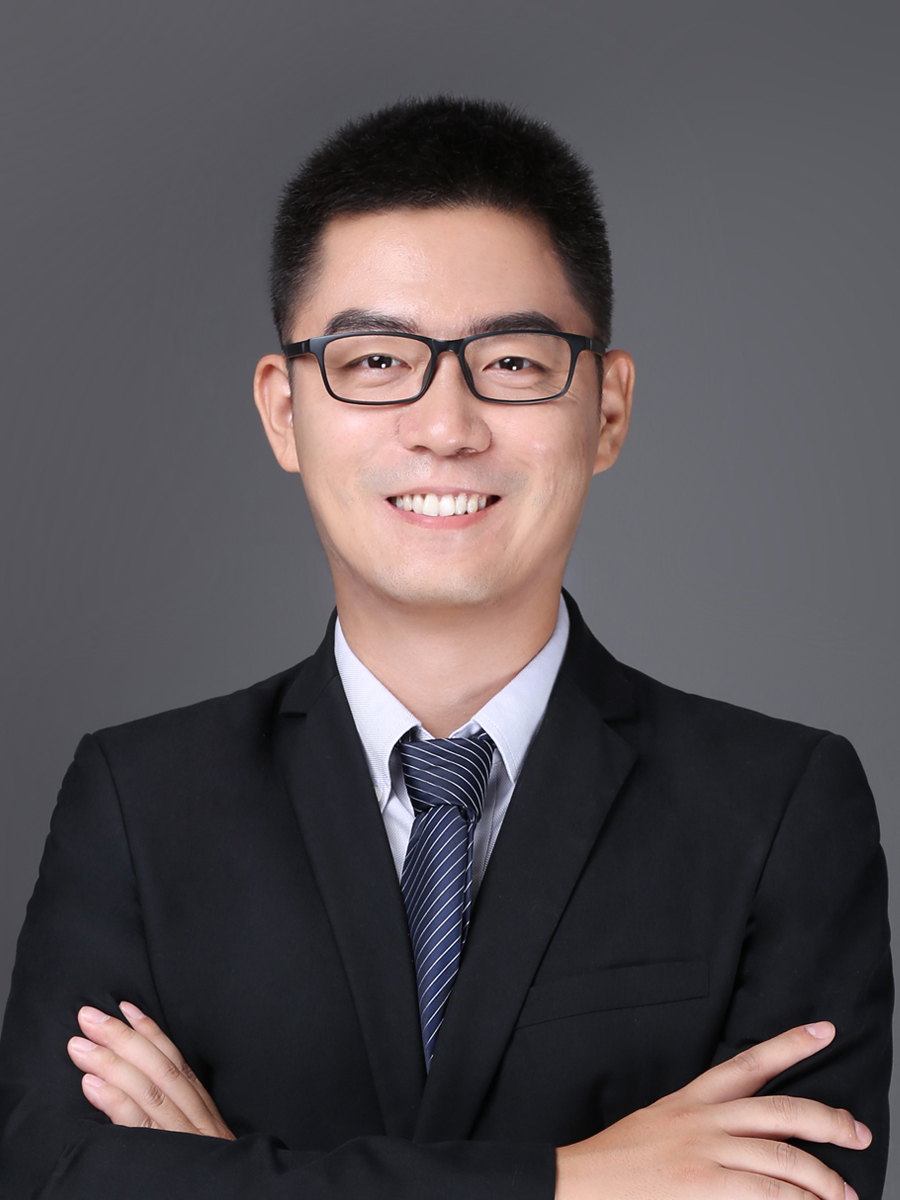}}]
{Boxin Shi} (Senior Member, IEEE) received the B.E. degree from the Beijing University of Posts and Telecommunications, the M.E. degree from Peking University, and the Ph.D. degree from the University of Tokyo, in 2007, 2010, and 2013. He is currently a Boya Young Fellow Associate Professor (with tenure) and Research Professor at Peking University, where he leads the Camera Intelligence Lab. Before joining PKU, he did research with MIT Media Lab, Singapore University of Technology and Design, Nanyang Technological University, National Institute of Advanced Industrial Science and Technology, from 2013 to 2017. His papers were awarded as Best Paper, Runners-Up at CVPR 2024, ICCP 2015 and selected as Best Paper candidate at ICCV 2015. He is an associate editor of TPAMI/IJCV and an area chair of CVPR/ICCV/ECCV. He is a senior member of IEEE.
\end{IEEEbiography}

\end{document}